\documentclass[10pt,twocolumn,letterpaper]{article}

\usepackage{cvpr}              % To produce the CAMERA-READY version
\makeatletter
\let\ESO@isMEMOIR\undefined
\let\AtPageUpperLeft\undefined

\let\AtTextUpperLeft\undefined

\let\ESO@HookI\undefined 
\let\ESO@HookII\undefined 
\let\ESO@HookIII\undefined 
\let\AddToShipoutPicture\undefined 
 
\let\ESO@yoffsetI\undefined 
\let\ESO@yoffsetII\undefined 
\makeatother

\usepackage{pdfpages}
\usepackage{times}
\usepackage{epsfig}
\usepackage{graphicx}
\usepackage{amsmath}
\usepackage{amssymb}
\usepackage{algorithm}
\usepackage{algpseudocode}

\usepackage{xcolor}         % colors
\usepackage{multicol, multirow}
\usepackage{booktabs}       % professional-quality tables
\usepackage{shadowtext}
\usepackage{natbib}        % for checkmark and crossmark
\usepackage{color,xcolor,colortbl}
\usepackage{empheq}
\usepackage{amsmath,amssymb,bm}

\renewcommand{\eqref}[1]{Eq.~(\ref{#1})}
\newcommand{\tabref}[1]{Table~\ref{#1}}
\newcommand{\secref}[1]{Sec.~\ref{#1}}
\newcommand{\figref}[1]{Fig.~\ref{#1}}

\newcommand{\vect}[1]{\mathbf{#1}}      % 向量
\newcommand{\mat}[1]{\mathbf{#1}}       % 矩阵
\newcommand{\ten}[1]{\mathcal{#1}}      % 张量
\newcommand{\set}[1]{\mathbb{#1}}      % 集合
\definecolor{cvprblue}{rgb}{0.21,0.49,0.74}
\usepackage[pagebackref,breaklinks,colorlinks,allcolors=cvprblue]{hyperref}
\definecolor{lightgray}{gray}{0.95}
\definecolor{color3}{gray}{0.95}
\definecolor{rouse}{rgb}{0.981,0.961,0.941}

\def\paperID{44781} % *** Enter the Paper ID here
\def\confName{CVPR}
\def\confYear{2026}

\title{LRDUN: A Low-Rank Deep Unfolding Network for Efficient Spectral Compressive Imaging}

\author{
	He Huang \quad Yujun Guo \quad Wei He\thanks{Corresponding author.} \\
	LIESMARS, Wuhan University, Wuhan, China\\
	{\tt\small huang\_he@whu.edu.cn} \quad {\tt\small yujunguo@whu.edu.cn} \quad {\tt\small weihe1990@whu.edu.cn}
}

\begin{document}
\maketitle
\begin{abstract}
	% "Deep unfolding networks (DUNs) have achieved remarkable success, emerging as the mainstream paradigm for spectral compressive imaging (SCI) reconstruction. However, most existing DUNs are predicated on the full-HSI imaging model, where each iteration alternates between a physics-based data-fidelity step and a deep prior refinement. Despite their effectiveness, these methods inherently suffer from severe ill-posedness and heavy computational burdens, as every stage mandates the reconstruction of the entire high-dimensional HSI directly from a single 2D coded measurement."

	Deep unfolding networks (DUNs) have achieved remarkable success and become the mainstream paradigm for spectral compressive imaging (SCI) reconstruction. 
	Existing DUNs are derived from full-HSI imaging models, where each stage operates directly on the high-dimensional HSI, refining the entire data cube based on the single 2D coded measurement. However, this paradigm leads to computational redundancy and suffers from the ill-posed nature of mapping 2D residuals back to 3D space of HSI.
	In this paper, we propose two novel imaging models corresponding to the spectral basis and subspace image by explicitly integrating low-rank (LR) decomposition with the sensing model. 
	Compared to recovering the full HSI, estimating these compact low-dimensional components significantly mitigates the ill-posedness. Building upon these novel models, we develop the Low-Rank Deep Unfolding Network (LRDUN), which jointly solves the two subproblems within an unfolded proximal gradient descent (PGD) framework. 
	Furthermore, we introduce a Generalized Feature Unfolding Mechanism (GFUM) that decouples the physical rank in the data-fidelity term from the feature dimensionality in the prior module, enhancing the representational capacity and flexibility of the network. Extensive experiments on simulated and real datasets demonstrate that the proposed LRDUN achieves state-of-the-art (SOTA) reconstruction quality with significantly reduced computational cost. 
	% The code is available at \url{https://github.com/huang-he99/LRDUN}.
	Code is available at \url{https://github.com/huang-he99/LRDUN}.

\end{abstract}
%%%%%%%%% BODY TEXT
\section{Introduction}

Hyperspectral images (HSI) capture rich spectral information across continuous wavelength bands, enabling numerous applications in remote sensing~\cite{hong2021interpretable,wang2025hypersigma}, medical imaging~\cite{backmanDetectionPreinvasiveCancer2000,luMedicalHyperspectralImaging2014}, and environmental monitoring~\cite{andrew2008role,rajabi2024hyperspectral}.
However, conventional HSI acquisition is time-consuming and data-intensive, hindering its applicability in real-time or large-scale scenarios.
Spectral compressive imaging (SCI)~\cite{yuan2021snapshot}, particularly snapshot-based systems such as coded aperture snapshot spectral imaging (CASSI)~\cite{wagadarikarSingleDisperserDesign2008,gehmSingleshotCompressiveSpectral2007,lin2014spatial,lin2014dual}, offers a compelling solution by directly capturing compressed measurements. This approach shifts the burden from hardware % -intensive 
acquisition to computational reconstruction. Nevertheless, recovering high-fidelity HSI from such heavily compressed data remains a severely ill-posed inverse problem, posing a fundamental challenge in SCI problem.

\begin{figure}[tbp]
    \centering
    \includegraphics[width=0.47\textwidth]{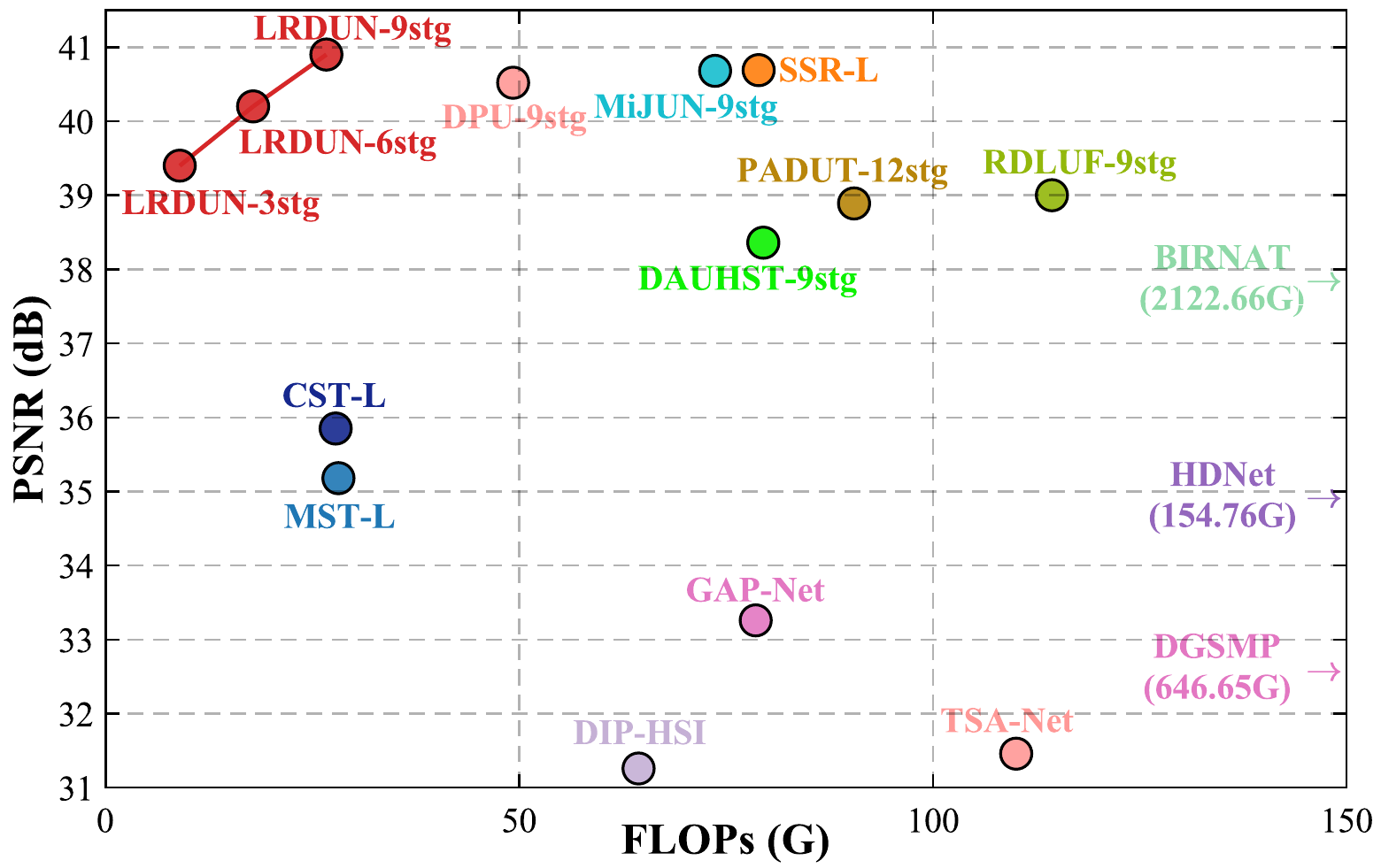}
    \vspace{-3mm}
    \caption{Performance (PSNR) vs. Efficiency (FLOPs). 
    LRDUN (red) achieves a superior accuracy-efficiency trade-off, 
    attaining competitive PSNR with significantly lower computational cost than SOTA methods.}
    \label{fig:psnrvsflops}
    \vspace{-5mm}
\end{figure}

To address this challenge,
% various reconstruction algorithms have been developed, 
% which can be 
% broadly categorized into model-based and deep learning-based methods. 
various model-based and deep learning-based algorithms have been developed.
Model-based methods~\cite{kittleMultiframeImageEstimation2010,yuanGeneralizedAlternatingProjection2016,wangAdaptiveNonlocalSparse2017,liuRankMinimizationSnapshot2019,zhangComputationalHyperspectralImaging2019} exploit handcrafted priors such as total variation (TV) \cite{yuanGeneralizedAlternatingProjection2016}, sparsity \cite{kittleMultiframeImageEstimation2010} and low-rank (LR) \cite{liuRankMinimizationSnapshot2019} to regularize the solution.
% Although these methods provide interpretability and theoretical guarantees, 
While interpretable and theoretically grounded,
these methods suffer from high computational cost and limited performance.

% To address this, various reconstruction algorithms have been developed, broadly categorized into model-based and deep learning-based methods. Model-based methods~\cite{kittleMultiframeImageEstimation2010,yuanGeneralizedAlternatingProjection2016,wangAdaptiveNonlocalSparse2017,liuRankMinimizationSnapshot2019,zhangComputationalHyperspectralImaging2019} rely on handcrafted priors such as total variation (TV) \cite{yuanGeneralizedAlternatingProjection2016}, sparsity \cite{kittleMultiframeImageEstimation2010} and low-rank \cite{liuRankMinimizationSnapshot2019} to constrain the solution space. Although they provide interpretability and theoretical guarantees, their performance is often limited by computational inefficiency and weak generalization under complex degradations.

With the rise of deep learning, three major classes of learning-based methods have emerged: end-to-end (E2E) networks, plug-and-play (PnP) frameworks, and deep unfolding networks (DUNs).
E2E methods~\cite{miaoLambdaNetReconstructHyperspectral2019,mengEndtoEndLowCost2020,huHDNetHighresolutionDualdomain2022,caiMaskguidedSpectralwiseTransformer2022,caiMSTMultistageSpectralwise2022,wangS2TransformerMaskAwareHyperspectral2025,mengSp3ctralMambaPhysicsDrivenJoint2025} directly learn the mapping from measurements to HSI, achieving high reconstruction accuracy but at the cost of requiring extensive training data and suffering from limited generalization and interpretability.
PnP frameworks~\cite{chanPlugandPlayADMMImage2017,qiaoSnapshotSpatialTemporal2020,yuanPlugandPlayAlgorithmsLargeScale2020,mengSelfsupervisedNeuralNetworks2021,zhengDeepPlugandplayPriors2021,yuanPlugandPlayAlgorithmsVideo2022,wuAdaptiveDeepPnP2023,xiePlugandPlayPriorsMultiShot2023} embed pre-trained denoisers into optimization algorithms, incorporating deep priors without retraining. However, the fixed denoiser limits their adaptability to hyperspectral characteristics.
In contrast, DUNs~\cite{wangHyperspectralImageReconstruction2019,wangDNUDeepNonLocal2020,mengDeepUnfoldingSnapshot2023,maDeepTensorADMMNet2019,huangDeepGaussianScale2021,zhangHerosNetHyperspectralExplicable2022,dongResidualDegradationLearning2023,liPixelAdaptiveDeep2023,huMAUNMemoryAugmentedDeep2024,zhangDualPriorUnfolding2024,zhangImprovingSpectralSnapshot2024,wuLatentDiffusionPrior2025,qinDetailMattersMambaInspired2025} unroll iterative algorithms into trainable neural networks, enabling end-to-end optimization while preserving algorithmic interpretability. With these merits, DUNs have demonstrated remarkable success and become a leading direction in SCI reconstruction.
% major research focus in this field.
% Benefiting from this balance between interpretability and learning capacity, DUNs have achieved remarkable success in SCI reconstruction and become a dominant research paradigm in the field.

Existing DUNs are built upon the full-HSI imaging model, alternating between data-fidelity enforcement and deep prior refinement.
However, this paradigm suffers from a fundamental bottleneck: the huge dimensionality gap compels the solver to infer numerous unknowns from limited measurements, leading to severe ill-posedness at each unfolding stage. Moreover, directly processing the high-dimensional data cube incurs heavy computational burdens, limiting the efficiency and scalability of the network.

To this end, we first derive two novel imaging models corresponding to the spectral basis and subspace image by explicitly integrating LR decomposition with the SCI sensing model. 
Specifically, the spectral basis captures global spectral correlations and material-dependent signatures, 
while the subspace images encode high-frequency spatial structures and local spectral sparsity. Crucially, this paradigm reformulates the high-dimensional reconstruction into tractable low-dimensional subproblems, significantly mitigating the inherent ill-posedness. 
Building upon these models, we develop the Low-Rank Deep Unfolding Network (LRDUN), a principled and efficient framework that jointly solves the two subproblems within an unfolded proximal gradient descent (PGD) optimization scheme. 
This yields an end-to-end network that achieves an elegant balance between interpretability, efficiency and fidelity.
Moreover, 
% a Generalized Feature Unfolding Mechanism (GFUM) is introduced to relax the coupling between the physical rank and the feature dimensionality, enabling both the data-fidelity and prior modules to operate in higher-dimensional feature spaces with adjustable complexity. This design substantially enhances the representational richness and adaptability of the network.
we introduce a Generalized Feature Unfolding Mechanism (GFUM) to relax the coupling between the physical rank and feature dimensionality. This mechanism empowers both the data-fidelity and prior modules to operate in higher-dimensional feature spaces, substantially enhancing the network's representational capacity and adaptability.
By bridging physics-based LR modeling and data-driven learning, LRDUN 
establishes a new paradigm for interpretable and efficient SCI reconstruction.
In summary, our main contributions are as follows:
\begin{itemize}
	\item We derive two novel and low-dimensional imaging models corresponding to the spectral basis and subspace image, which reformulate the ill-posed SCI problem into two efficiently solvable subproblems.
	\item We develop LRDUN that jointly solves the decomposed subproblems via an unfolded PGD scheme, coupling physical insight with learnable networks.
	\item We introduce a flexible GFUM that decouples the physical rank from the network's feature dimension, enhancing the representational capacity and flexibility.
	\item LRDUN achieves state-of-the-art reconstruction quality with significantly reduced computational complexity, establishing a superior accuracy-efficiency trade-off.
\end{itemize}

\noindent
\textbf{Notations.}
% We follow the tensor notation in \cite{koldaTensorDecompositionsApplications2009},
% the tensor, matrix and vector are represented as Euler script letters, \textit{i.e.} $\ten{X}$, boldface capital letter, \textit{i.e.} $\mat{A}$ and boldface lowercase letter, \textit{i.e.} $\vect{x}$ respectively.
% For a $N$-order tensor $\ten{X}\in\set{R}^{I_1\times I_2\times\cdots \times I_N}$,
% the mode-$n$ unfolding operator is denoted as $\mat{X}_{(n)}\in\set{R}^{I_n \times  {I_1 \cdots I_{n-1} I_{n+1} \cdots I_N}}$.
% We have $\text{fold}_n(\mat{X}_{(n)})=\ten{X}$, in which $\text{fold}_n$ is the inverse operator of unfolding operator. 
% The mode-$n$ product of a tensor $\ten{X}\in\set{R}^{I_1\times I_2\times\cdots \times I_N}$ and a matrix
% $\mat{A}\in\set{R}^{J_n\times I_n}$ is defined as $\ten{Y} = \ten{X} \times_n
% \mat{A}$, where $\ten{Y}\in\set{R}^{I_1\times I_2\times\cdots \times J_n}$ and $\ten{X}\times_n \mat{A} = \text{fold}_n(\mat{A} \mat{X}_{(n)})$. $\odot$ denotes Hadamard product. $\otimes$ denotes Kronecker product. $\mathrm{vec}(\cdot)$ denotes vectorization operation in a consistent order (mode-1 to mode-$N$).
We follow the tensor notations in \cite{koldaTensorDecompositionsApplications2009}.
Tensors, matrices, and vectors are denoted by Euler script letters (e.g., $\ten{X}$), boldface capital letters (e.g., $\mat{A}$), and boldface lowercase letters (e.g., $\vect{x}$), respectively.
For an $N$-th order tensor $\ten{X}\in\set{R}^{I_1\times I_2\times\cdots\times I_N}$, its mode-$n$ unfolding is denoted by $\mat{X}_{n}\in\set{R}^{I_n\times (I_1\cdots I_{n-1}I_{n+1}\cdots I_N)}$, and the inverse operation is $\text{fold}_n(\mat{X}_{n})=\ten{X}$.
The mode-$n$ product between $\ten{X}$ and a matrix $\mat{A}\in\set{R}^{J_n\times I_n}$ is defined as $\ten{Y}=\ten{X}\times_n\mat{A}$, where $\ten{Y}\in\set{R}^{I_1\times\cdots\times J_n\times\cdots\times I_N}$ and $\ten{X}\times_n\mat{A}=\text{fold}_n(\mat{A}\mat{X}_{n})$.
$\odot$ and $\otimes$ denote the Hadamard and Kronecker products, respectively.
$\mathrm{vec}(\cdot)$ denotes the vectorization operator applied in a consistent order (mode-1 to mode-$N$).

\section{Related Work}
% In this section, we first illustrate the spectral low-rank property of HSI, and subsequently introduce the development of deep-learningbased methods for HSI reconstruction.
In this section, we review two key areas relevant to our work: the development of  DUNs for SCI and utilization of the spectral LR property in HSI processing.

\subsection{DUNs for SCI}
Conventional model-based methods typically cast SCI reconstruction as an MAP optimization problem, solved via iterative algorithms (e.g. HQS \cite{nikolovaAnalysisHalfQuadraticMinimization2005}, ADMM \cite{boydDistributedOptimizationStatistical2011}, GAP \cite{liaoGeneralizedAlternatingProjection2014}, PGD \cite{beckFastIterativeShrinkageThresholding2009}) that alternate between data-fidelity and prior regularization steps.
DUNs extend this paradigm by unfolding the iterative process into a series of learnable stages, each comprising a linear update and a trainable prior module.
Early DUNs, such as DSSP \cite{wangHyperspectralImageReconstruction2019}, GAP-Net \cite{mengDeepUnfoldingSnapshot2023}, DGSMP \cite{huangDeepGaussianScale2021} and DNU \cite{wangDNUDeepNonLocal2020}, primarily employed CNNs to model local correlations.
The first Transformer-based DUN, DAUHST \cite{caiDegradationawareUnfoldingHalfshuffle2022}, customized a Half-Shuffle Transformer (HST) to capture local content and nonlocal dependencies, followed by variants such as PADUT \cite{liPixelAdaptiveDeep2023}, RDLUF \cite{dongResidualDegradationLearning2023}, SSR \cite{zhangImprovingSpectralSnapshot2024} and DPU \cite{zhangDualPriorUnfolding2024} for enhanced spectral and spatial prior modeling.
Recently,
DUNs have begun to incorporate diffusion priors (e.g., LADE-DUN \cite{wuLatentDiffusionPrior2025}) and Mambas (e.g., MiJUN \cite{qinDetailMattersMambaInspired2025}) to further improve reconstruction fidelity and efficiency. However,
% Despite their effectiveness, existing DUNs 忽视了HSI的固有低秩特性，另外suffer from high computational costs due to full HSI cube processing at each stage.
existing DUNs operate within the full-HSI solution space, which not only imposes high computational overhead but also necessitates back-projecting 2D residuals to the 3D space of HSI, creating a massive dimensionality gap and severe ill-posedness.

\begin{figure*}[tbp]
    \centering
    \includegraphics[width=0.98\textwidth]{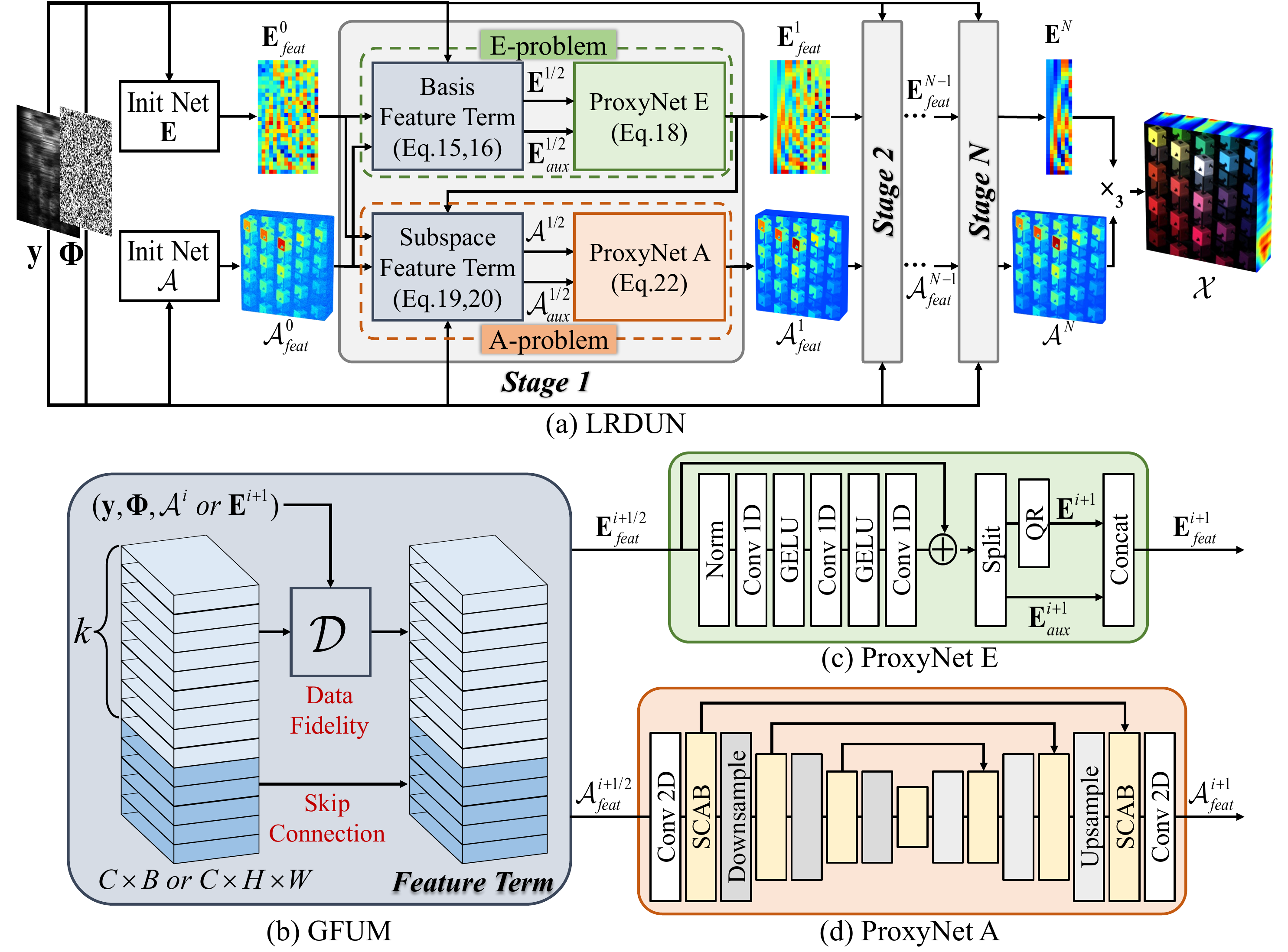}
    \vspace{-3mm}
    \caption{
    %     Overall architecture of the proposed LRDUN. 
    % Our network unfolds the alternating PGD optimization algorithm into $N$ stages. 
    % Each stage contains two core modules: the E-problem module for updating the spectral basis and the A-problem module for updating the subspace images. 
    % As illustrated, both modules alternate between a data-fidelity term (Feature Term) and a learnable prior (ProxyNet). 
    % An Initial Network (Init Net) is used to initialize $\mathcal{A}_{feat}^0$ and $E_{feat}^0$ from the compressed measurement $y$.
    % Overall architecture of LRDUN. The network unfolds an optimization into $N$ stages. Each stage alternates between solving an E-problem (spectral basis) 
    % and an A-problem (subspace images), each comprising a data-fidelity term (Feature Term) and a learnable prior (ProxyNet).
    Overall architecture of LRDUN. The network unfolds an optimization into N stages. Each stage alternates between solving an E-problem (spectral basis feature) and an A-problem (subspace images feature), each comprising a data-fidelity feature term via (b) GFUM and a learnable (c) ProxyNet E or (d) ProxyNet A.
    }
    \label{fig:arch}
    \vspace{-3mm}
\end{figure*}

\subsection{Spectral LR Property of HSI}

HSIs exhibit strong spectral correlations, resulting in a natural LR structure widely exploited in HSI denoising \cite{zhangHyperspectralImageRestoration2013}, super-resolution \cite{xueSpatialspectralStructuredSparse2021}, and inpainting \cite{zhuangFastHyperspectralImage2018}. In SCI reconstruction, early methods utilized this property through nuclear norm minimization \cite{liuRankMinimizationSnapshot2019, zhangComputationalHyperspectralImaging2019}, subspace learning \cite{chenHyperspectralImageCompressive2020}, or tensor decomposition \cite{xuHyperspectralComputationalImaging2020}. Later approaches combined LR priors with deep learning, often within the PnP framework \cite{chenCombiningLowrankDeep2023, chenHyperspectralCompressiveSnapshot2024, chenFlexDLDDeepLowrank2024}, employing deep image priors (DIP) to estimate LR components. However, these approaches often suffer from slow convergence, manual parameter tuning, and high computational overhead. 
Attempts to integrate LR priors into DUNs have also been limited. For instance, TLPLN \cite{zhangLearningTensorLowRank2021} integrated tensor CP decomposition into DUNs
but still processes the full HSI, failing to tightly couple the prior with the sensing physics. He et al. \cite{heInterpretableFlexibleFusion2024} proposed a subspace distillation prior (SP) as a post-processing module for DUNs, but its performance is constrained by existing DUNs. Fundamentally, existing methods treat the LR property as an auxiliary regularizer or post-processor, without altering the full-HSI imaging model in the data-fidelity term. Consequently, the reconstruction remains burdened by the high dimensionality of the target variable. In contrast, our method explicitly reformulates the sensing model itself via LR decomposition, directly solving for low-dimensional components to mitigate the ill-posedness.

\section{Imaging Model of CASSI}
A typical CASSI system spatially modulates the scene through a coded aperture, introduces wavelength-dependent dispersion through a dispersive element, and finally records the modulated light on a focal plane array (FPA) detector.

Consider an HSI $\ten{X} \in \mathbb{R}^{H \times W \times B}$, where $\mat{X}_{b} \in \mathbb{R}^{H \times W}$ denotes its 2D spatial image slice at wavelength $\lambda_{b}$. This scene is first spatially modulated by a coded aperture $\mat{M} \in \mathbb{R}^{H \times W}$, yielding an intermediate result:
\begin{align}
	\label{eq:coded_aperture}
	\mat{X}'_{b} = \mat{M} \odot \mat{X}_{b}.
\end{align}
Subsequently, a dispersive element shifts the modulated image $\mat{X}'_{b}$ along a spatial dimension (e.g., horizontally) by an offset $d(\lambda_{b})$ proportional to the wavelength. The detector integrates all modulated and shifted spectral components, producing the final 2D compressed measurement $\mat{Y} \in \mathbb{R}^{H \times W^\prime}$, where $W^\prime = W + d(B-1)$ is the effective width after compression:
\begin{align}
	\label{eq:image_model}
	\mat{Y}(h, w) = \sum_{b=1}^{B} \mat{X}'_{b}(h,w+d(\lambda_{b})).
\end{align}
Incorporating the measurement noise, the entire imaging model \eqref{eq:image_model} can be expressed in matrix-vector form as:
\begin{align}
	\label{eq:cassi_model}
	\vect{y} = \mat{\Phi} \vect{x} + \vect{n},
\end{align}
where $\vect{y} = \mathrm{vec}(\mat{Y})$ and $\vect{x} = \mathrm{vec}(\ten{X})$ are the vectorized form of the compressed measurement and the target HSI, respectively; $\mat{\Phi} \in \mathbb{R}^{HW' \times HWB}$ denotes the system sensing matrix, and $\vect{n}$ denotes the measurement noise. The goal of SCI is to reconstruct the original HSI $\ten{X}$ from the compressed 2D measurement $\mat{Y}$.

\section{Method}

In this section, we present the proposed LRDUN. We first reformulate the CASSI imaging model by integrating the HSI's low-rank property, deriving novel spectral basis and spatial subspace imaging models (\secref{sec:basis_subspace}). Based on these, we formulate a joint optimization problem solved via an alternating PGD algorithm (\secref{subsec:model_optimization}). We then unroll this algorithm into an end-to-end network, introducing a Generalized Feature Unfolding Mechanism (GFUM) to enhance representational capacity (\secref{subsec:GFUM}). Finally, we present the overall N-stage network architecture (\secref{subsec:over}) and detail the structures of ProxyNets (\secref{subsec:proxynet}).

\subsection{Basis and Subspace Imaging Models}
\label{sec:basis_subspace}
In this subsection, we integrate the original CASSI imaging model with the LR decomposition of HSI, further leading to two novel imaging models: the spectral basis and spatial subspace imaging models.

Firstly, HSIs often exhibit strong spectral correlations, allowing them to be effectively represented in a low-dimensional subspace \cite{heNonlocalMeetsGlobal2020, heSpectrumAwareTransferableArchitecture2022}. We have the following formulation:
\begin{align}
	\label{eq:low_rank_decomposition}
	\ten{X} = \ten{A} \times_3 \mat{E},
\end{align}
where $\mat{E} \in \mathbb{R}^{B \times k}$ is the spectral basis, and $\ten{A} \in \mathbb{R}^{H \times W \times k}$ represents the corresponding subspace images. Here, $k$ is the physical rank of the HSI, typically satisfying $k \ll B$. The corresponding matrix form of \eqref{eq:low_rank_decomposition} is given by:
\begin{align}
	\label{eq:low_rank_matrix}
	\mat{X}_{(3)} = \mat{E} \mat{A}_{(3)}.
\end{align}
For notational simplicity, we denote $\mat{A}=\mat{A}_{(3)}^T$. Thus, the imaging model \eqref{eq:cassi_model} can be rewritten as:
\begin{align}
	\vect{y} = \mat{\Phi}  \mathrm{vec}(\mat{X}_{(3)}^T) + \vect{n} = \mat{\Phi}  \mathrm{vec}(\mat{A} \mat{E}^T) + \vect{n}.
\end{align}
Notice a key property of vectorization: $\mathrm{vec}(\mat{U}\mat{V}\mat{W}) = (\mat{W}^T\otimes\mat{U}) \mathrm{vec}(\mat{V})$. By letting $(\mat{U}, \mat{V}, \mat{W}) = (\mat{A}, \mat{E}^T, \mat{I}_{B})$  and $(\mat{U}, \mat{V}, \mat{W}) = (\mat{I}_{HW}, \mat{A}, \mat{E}^T)$, we can obtain two imaging models regarding spectral basis and spatial subspace:
% \textcolor{red}{
% 	\begin{empheq}[left=\empheqlbrace]{align}
% 		\vect{y} &= \mat{\Phi}_{\mat{A}}  \mathrm{vec}(\mat{E}^{T}) + \vect{n}, \label{eq:basis} \\
% 		\vect{y} &= \mat{\Phi}_{\mat{E}}  \mathrm{vec}(\mat{A}) + \vect{n}, \label{eq:subspace}
% 	\end{empheq}
% }
% where $\mat{\Phi}_{\mat{A}} = \mat{\Phi} (\mat{I}_{B} \otimes \mat{A})$ and $\mat{\Phi}_{\mat{E}} = \mat{\Phi} (\mat{E} \otimes \mat{I}_{HW})$ are the sensing matrices for the spectral basis and spatial subspace imaging models, respectively.
\textcolor{red}{
	\begin{empheq}[left=\empheqlbrace]{align}
		\vect{y} &= \mat{\Phi}_{\mat{A}}  \vect{e} + \vect{n}, \label{eq:basis} \\
		\vect{y} &= \mat{\Phi}_{\mat{E}}  \vect{a} + \vect{n}, \label{eq:subspace}
	\end{empheq}
}
where $\vect{e}=\mathrm{vec}(\mat{E}^{T})$, $\vect{a}=\mathrm{vec}(\mat{A})$; $\mat{\Phi}_{\mat{A}} = \mat{\Phi} (\mat{I}_{B} \otimes \mat{A})$ and $\mat{\Phi}_{\mat{E}} = \mat{\Phi} (\mat{E} \otimes \mat{I}_{HW})$ are the sensing matrices for the spectral basis and spatial subspace imaging models, respectively.

While mathematically derived from the original system, the proposed basis (\eqref{eq:basis}) and subspace (\eqref{eq:subspace}) imaging models offer a crucial advantage in optimization. The original model (\eqref{eq:cassi_model}) requires solving for $N = H \times W \times B$ unknowns. In contrast, our dual representation reduces the problem to estimating $\mat{E} \in \set{R}^{B \times k}$ and $\mat{A} \in \set{R}^{HW \times k}$. Since the physical rank $k \ll B$, the number of unknowns is drastically reduced. This dimensionality reduction significantly mitigates the severe ill-posedness inherent in the original SCI inverse problem, converting the challenging reconstruction of a full data cube into the estimation of compact low-dimensional components.

% The two imaging models in \eqref{eq:basis} and \eqref{eq:subspace} are mathematically equivalent but offer distinct physical insights.
% The basis form \eqref{eq:basis} estimates the spectral bases $\mat{E}$ that capture material-dependent spectral signatures, with the sensing matrix $\mat{\Phi}_{\mat{A}}$ embedding spatial subspace information $\mat{A}$.
% Conversely, the subspace form \eqref{eq:subspace} reconstructs the spatial coefficients $\mat{A}$ describing scene structures, where $\mat{\Phi}_{\mat{E}}$ encodes the spectral prior $\mat{E}$.
% This dual representation decouples spectral and spatial components in the sensing process, reducing the original reconstruction of $\ten{X} \in \mathbb{R}^{H \times W \times B}$ to two compact sub-problems: estimating $\mat{E} \in \mathbb{R}^{B \times k}$ and $\mat{A} \in \mathbb{R}^{H W \times k}$.
% Since $k \ll B$, both are substantially lower-dimensional, mitigating ill-posedness and enabling
% % an alternating optimization (AO) scheme that iteratively updates $\mat{E}$ and $\mat{A}$ via \eqref{eq:basis} and \eqref{eq:subspace}. 
% % This formulation provides 
% a concise and flexible foundation for subsequent reconstruction networks and algorithmic designs.

\subsection{Model Optimization}
\label{subsec:model_optimization}
Based on the imaging models in \eqref{eq:basis} and (\ref{eq:subspace}), we formulate the SCI as the following optimization problem:
% \begin{align}
% 	\label{eq:optimization}
% 	\min_{\vect{e}, \vect{a}} \frac{1}{2} \| \vect{y} - \mat{\Phi}_{\mat{A}} \vect{e} \|_2^2 + \frac{1}{2} \| \vect{y} - \mat{\Phi}_{\mat{E}}  \vect{a}  \|_2^2 + \lambda_{\vect{e}} \mathcal{R}_{\vect{e}}(\vect{e}) + \lambda_{\vect{a}} \mathcal{R}_{\vect{a}}(\vect{a}),
% \end{align}
\begin{align}
	\label{eq:optimization}
	\min_{\vect{e}, \vect{a}} \frac{1}{2} \| \vect{y} - \mat{\Phi}_{\mat{A}} \vect{e} \|_2^2 + \frac{1}{2} \| \vect{y} - \mat{\Phi}_{\mat{E}}  \vect{a}  \|_2^2 + \lambda_{\vect{e}} \mathcal{R}_{\vect{e}}(\vect{e}) + \lambda_{\vect{a}} \mathcal{R}_{\vect{a}}(\vect{a}),
\end{align}
where 
% vectorized forms are used for simplicity; 
$\mathcal{R}_{\vect{e}}(\cdot)$ and $\mathcal{R}_{\vect{a}}(\cdot)$ are regularization terms for spectral basis and spatial subspace, respectively; $\lambda_{\vect{e}}$ and $\lambda_{\vect{a}}$ are the corresponding trade-off parameters.
% Similar to \cite{fuKXNetModelDrivenDeep2022}, we adopt a proximal gradient descent (PGD) technique \cite{beckFastIterativeShrinkageThresholding2009, gregorLearningFastApproximations2010} to solve the \eqref{eq:optimization} by alternately updating $\vect{e}$ and $\vect{a}$: 
Following \cite{fuKXNetModelDrivenDeep2022}, we employ the proximal gradient descent (PGD) method \cite{beckFastIterativeShrinkageThresholding2009, gregorLearningFastApproximations2010} to solve \eqref{eq:optimization} by alternately updating $\vect{e}$ and $\vect{a}$.
% The resulting optimization procedure involves only simple and interpretable operators, making it readily unfoldable into corresponding network modules. 
The i-th iteration updates are given by:

\noindent
\textbf{E-problem:} The update of spectral basis $\vect{e}$ can be optimized by solving the following quadratic problem with fixed $\vect{a}^{i}$:
\begin{align}
	\label{eq:update_e}
	\vect{e}^{i+1} = \arg\min_{\vect{e}} \frac{1}{2} \| \vect{y} - \mat{\Phi}_{\mat{A}^{i}} \vect{e} \|_2^2 + \lambda_{\vect{e}} \mathcal{R}_{\vect{e}}(\vect{e}).
\end{align}
Using PGD, we first compute the GD step:
\begin{align}
	\label{eq:pgd_e}
	\vect{e}^{i+1/2} = \vect{e}^{i} - \rho_{\vect{e}} \mat{\Phi}_{\mat{A}^{i}}^{T} (\mat{\Phi}_{\mat{A}^{i}} \vect{e}^{i} - \vect{y}),
\end{align}
where $\rho_{\vect{e}}$ is the step size. Then, we apply the proximal operator associated with $\mathcal{R}_{\vect{e}}(\cdot)$:
\begin{align}
	\label{eq:prox_e}
	\vect{e}^{i+1} = \mathrm{prox}_{\lambda_{\vect{e}}\rho_{\vect{e}}, \mathcal{R}_{\vect{e}}}(\vect{e}^{i+1/2}).
\end{align}
\textbf{A-problem:} Similarly, the update of spatial subspace $\vect{a}$ is obtained by solving (where $\rho_{\vect{a}}$ is the step size):
% \begin{align}
%     \label{eq:update_a}
%     \vect{a}^{(i+1)} = \arg\min_{\vect{a}} 1/2 \| \vect{y} - \mat{\Phi}_{\vect{e}^{(i+1)}} \vect{a} \|_2^2 + \lambda_{\vect{a}} \mathcal{R}_{\vect{a}}(\vect{a}).
% \end{align}
% The PGD steps are given by:
\begin{align}
	\vect{a}^{i+1/2} & = \vect{a}^{i} - \rho_{\vect{a}} \mat{\Phi}_{\mat{E}^{i+1}}^{T} (\mat{\Phi}_{\mat{E}^{i+1}} \vect{a}^{i} - \vect{y}), \label{eq:pgd_a} \\
	\vect{a}^{i+1}   & = \mathrm{prox}_{\lambda_{\vect{a}}\rho_{\vect{a}}, \mathcal{R}_{\vect{a}}}(\vect{a}^{i+1/2}). \label{eq:prox_a}
\end{align}

\subsection{Generalized Feature Unfolding Mechanism}
\label{subsec:GFUM}
The alternating optimization algorithm derived in \secref{subsec:model_optimization} (Eq. (\ref{eq:pgd_e}-\ref{eq:prox_a})) involves only simple linear and proximal operators, making it well-suited for unrolling into a deep network. Specifically, each iteration can be mapped to one network stage, where the GD step (Eq. (\ref{eq:pgd_e}) and (\ref{eq:pgd_a})) corresponds to physics-driven data-fidelity terms $(\ten{D}_{\mathrm{E}}, \ten{D}_{\mathrm{A}})$, and the proximal steps (Eq. (\ref{eq:prox_e}) and (\ref{eq:prox_a})) are implemented as learnable prior modules $(\mathrm{ProxyNet}_{\mathrm{E}}, \mathrm{ProxyNet}_{\mathrm{A}})$.

However, a key limitation arises in this straightforward unrolling process: an inherent coupling between the physical rank $k$ and the feature dimensionality of the ProxyNets. Since the data-fidelity terms are defined in the $k$-dim physical space, the input and output of the ProxyNets are also constrained to this $k$-dim manifold. This severely restricts the representational capacity of the ProxyNets, hindering the ability to learn complex spatial-spectral dependencies.

To overcome this bottleneck, we introduce a Generalized Feature Unfolding Mechanism (GFUM) that empowers the data-fidelity terms and prior modules to operate in higher-dim feature spaces. 
As illustrated in \figref{fig:arch}(b), we elevate the $k$-dim optimization variables (e.g., $\ten{A}$ or $\mat{E}$) into the $C$-dim feature representations ($\ten{A}_{feat}$ or $\mat{E}_{feat}$) with $C \ge k$. The resulting features are conceptually partitioned into two distinct parts: a $k$-dim physical component that enforces data fidelity and the complementary $(C-k)$-dim auxiliary component as an information carrier, which is preserved in the data-fidelity term and assists in the physical component's refinement within the ProxyNet. 
Specifically, the update of each variable ($\mat{E}$ and $\ten{A}$) under the GFUM can be expressed as follows:

\noindent
\textbf{E-problem:}
\vspace{-2mm}
\begin{align}
        & \mat{E}^{i}             = \mat{E}_{feat}^{i}(:k) \rightarrow \mat{E}^{i+1/2}         = \ten{D}_{\mathrm{E}}(\mat{E}^{i}, \vect{y}, \mat{\Phi}, \ten{A}^{i}), \\
        & \mat{E}_{aux}^{i}       = \mat{E}_{feat}^{i}(k:)  \rightarrow  \mat{E}_{aux}^{i+1/2} = \mat{E}_{aux}^{i},                                                \\
        & \mat{E}_{feat}^{i+1/2}  = [\mat{E}^{i+1/2}; \mat{E}_{aux}^{i+1/2}], \label{eq:fdm_concat_e}                                                                    \\
        & \mat{E}_{feat}^{i+1}    = \mathrm{ProxyNet}_{\mathrm{E}}(\mat{E}_{feat}^{i+1/2}), \label{eq:fdm_prox_e}
\end{align}

\vspace{-2mm}
\noindent
\textbf{A-problem:}
\vspace{-2mm}
\begin{align}
	 & \ten{A}^{i}             = \ten{A}_{feat}^{i}(:k) \rightarrow \ten{A}^{i+1/2}         = \ten{D}_{\mathrm{A}}(\ten{A}^{i}, \vect{y}, \mat{\Phi}, \mat{E}^{i+1}), \\
	 & \ten{A}_{aux}^{i}       = \ten{A}_{feat}^{i}(k:)  \rightarrow  \ten{A}_{aux}^{i+1/2} = \ten{A}_{aux}^{i},                                                \\
	 & \ten{A}_{feat}^{i+1/2}  = [\ten{A}^{i+1/2}; \ten{A}_{aux}^{i+1/2}], \label{eq:fdm_concat}                                                                    \\
	 & \ten{A}_{feat}^{i+1}    = \mathrm{ProxyNet}_{\mathrm{A}}(\ten{A}_{feat}^{i+1/2}). \label{eq:fdm_prox}
\end{align}
% where $\ten{A}_{feat}$ is the $C$-dimensional feature output from the gradient descent step (i.e., Feature Term). A symmetric feature decoupling strategy is adopted for the update of the spectral basis $E$. 
% This symmetric mechanism allows the network to enjoy the flexibility and powerful representational capacity of high-dimensional feature learning while maintaining the physical interpretability of the low-rank optimization, thereby achieving a balance between physics-based modeling and data-driven representation.
Notably, despite its simplicity, the auxiliary component is crucial. It implicitly learns to carry diverse, spatially-varying information, such as proximal parameters, noise residuals, and reconstruction artifacts, or even encoding the physical CASSI mask, which is empirically validated in the subsequent visualization \secref{subsec:feat_visual}. 
% Compared with \cite{chen2023deep, chen2025self},  GFUM扩展到多变量的交替优化，并将aux视为信息具有物理意义的信息载体，而非单纯的feature channel。
Compared with \cite{chen2023deep, chen2025self}, GFUM is extended to the alternating optimization of multiple variables and treats the auxiliary component as a physical information carrier rather than just feature channels.
\begin{figure}[tbp]
    \centering
    \includegraphics[width=0.27\textwidth]{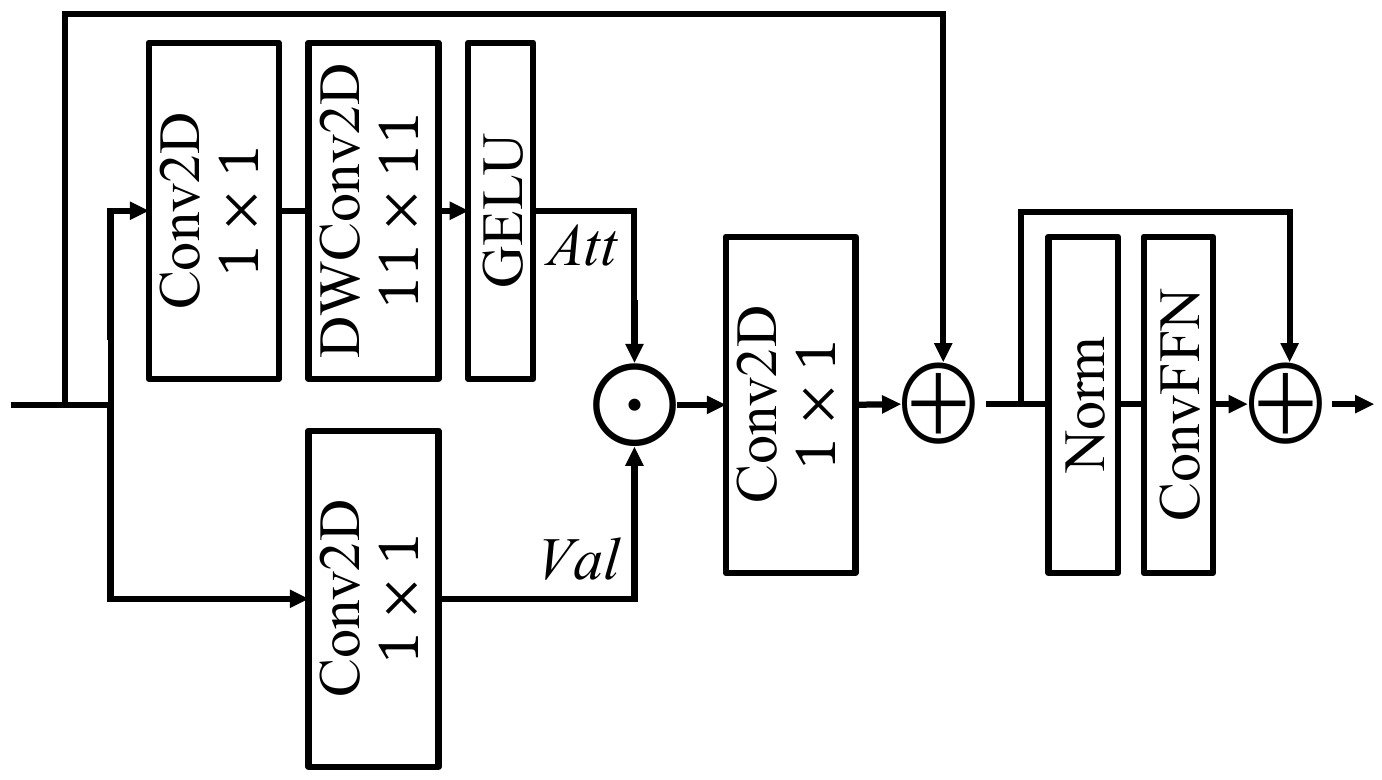}
    \vspace{-3mm}
    \caption{The architecture of the SCAB.}
    \label{fig:sab_detail}
    \vspace{-8mm}
\end{figure}

\begin{table*}[tbp]
	\renewcommand{\arraystretch}{0.7} % 调整行间距
	\centering
	\caption{The PSNR (upper entry in each cell) in dB and SSIM (lower entry in each cell) results of the test methods on 10 scenes.}
	\vspace{-3mm}
	\label{tab:simu}
	% 为了确保表格能适应页面宽度，可以使用 resizebox 或调整字体大小
	\scriptsize
	\resizebox{\textwidth}{!}{%
		% \small
		\begin{tabular}{ccccccccccccccc}
			\toprule[0.2em]
			\rowcolor{lightgray}
			{Algorithms}                   & {Reference}                   & {Params (M)}           & {FLOPs (G)}              & {S1}  & {S2}  & {S3}  & {S4}    & {S5}  & {S6}  & {S7}  & {S8}  & {S9}  & {S10} & {Avg} \\
			\midrule
			\multirow{2}{*}{TwIST}         & \multirow{2}{*}{TIP 2017}     & \multirow{2}{*}{-}     & \multirow{2}{*}{-}       & 25.16 & 23.02 & 21.40 & 30.19   & 21.41 & 20.95 & 22.20 & 21.82 & 22.42 & 22.67 & 23.12 \\
			                               &                               &                        &                          & 0.700 & 0.604 & 0.711 & 0.851   & 0.635 & 0.644 & 0.643 & 0.650 & 0.690 & 0.584 & 0.669 \\
			\midrule
			\multirow{2}{*}{GAP-TV }       & \multirow{2}{*}{ICIP 2016}    & \multirow{2}{*}{-}     & \multirow{2}{*}{-}       & 26.82 & 22.89 & 26.31 & 30.65   & 23.64 & 21.85 & 23.76 & 21.98 & 22.62 & 23.10 & 24.36 \\
			                               &                               &                        &                          & 0.754 & 0.610 & 0.802 & 0.852   & 0.703 & 0.663 & 0.688 & 0.655 & 0.682 & 0.584 & 0.690 \\
			\midrule
			\multirow{2}{*}{DeSCI}         & \multirow{2}{*}{TPAMI 2019}   & \multirow{2}{*}{-}     & \multirow{2}{*}{-}       & 27.13 & 23.04 & 26.62 & 34.96   & 23.94 & 22.38 & 24.45 & 22.03 & 24.56 & 23.59 & 25.27 \\
			                               &                               &                        &                          & 0.748 & 0.620 & 0.818 & 0.897   & 0.706 & 0.683 & 0.743 & 0.673 & 0.732 & 0.587 & 0.721 \\
			\midrule
			\multirow{2}{*}{$\lambda$-Net} & \multirow{2}{*}{ICCV 2019}    & \multirow{2}{*}{62.64} & \multirow{2}{*}{117.98}  & 30.10 & 28.49 & 27.73 & 37.01   & 26.19 & 28.64 & 26.47 & 26.09 & 27.50 & 27.13 & 28.53 \\
			                               &                               &                        &                          & 0.849 & 0.805 & 0.870 & 0.934   & 0.817 & 0.853 & 0.806 & 0.831 & 0.826 & 0.816 & 0.841 \\
			% \midrule
			% \multirow{2}{*}{HSSP}          & \multirow{2}{*}{CVPR 2019}    & \multirow{2}{*}{-}     & \multirow{2}{*}{-}       & 31.48 & 31.09 & 28.96 & 34.56   & 28.53 & 30.83 & 28.71 & 30.09 & 30.43 & 28.78 & 30.35 \\
			%                                &                               &                        &                          & 0.858 & 0.842 & 0.823 & 0.902   & 0.808 & 0.877 & 0.824 & 0.881 & 0.868 & 0.842 & 0.852 \\
			% \midrule
			% \multirow{2}{*}{DNU}           & \multirow{2}{*}{CVPR 2020}    & \multirow{2}{*}{1.19}  & \multirow{2}{*}{163.48}  & 31.72 & 31.13 & 29.99 & 35.34   & 29.03 & 30.87 & 28.99 & 30.13 & 31.03 & 29.14 & 30.74 \\
			%                                &                               &                        &                          & 0.863 & 0.846 & 0.845 & 0.908   & 0.833 & 0.887 & 0.839 & 0.885 & 0.876 & 0.849 & 0.863 \\
			\midrule
			\multirow{2}{*}{DIP-HSI}       & \multirow{2}{*}{ICCV 2021}    & \multirow{2}{*}{33.85} & \multirow{2}{*}{64.42}   & 32.68 & 27.26 & 31.30 & 40.54   & 29.79 & 30.39 & 28.18 & 29.44 & 34.51 & 28.51 & 31.26 \\
			                               &                               &                        &                          & 0.890 & 0.833 & 0.914 & 0.962   & 0.900 & 0.877 & 0.913 & 0.874 & 0.927 & 0.851 & 0.894 \\
			\midrule
			\multirow{2}{*}{TSA-Net}       & \multirow{2}{*}{ECCV 2020}    & \multirow{2}{*}{44.25} & \multirow{2}{*}{110.06}  & 32.03 & 31.00 & 32.25 & 39.19   & 29.39 & 31.44 & 30.32 & 29.35 & 30.01 & 29.59 & 31.46 \\
			                               &                               &                        &                          & 0.892 & 0.858 & 0.915 & 0.953   & 0.884 & 0.908 & 0.878 & 0.888 & 0.890 & 0.874 & 0.894 \\
			\midrule
			\multirow{2}{*}{DGSMP}         & \multirow{2}{*}{CVPR 2021}    & \multirow{2}{*}{3.76}  & \multirow{2}{*}{646.65}  & 33.26 & 32.09 & 33.06 & 40.54   & 28.86 & 33.08 & 30.74 & 31.55 & 31.66 & 31.44 & 32.63 \\
			                               &                               &                        &                          & 0.915 & 0.898 & 0.925 & 0.964   & 0.882 & 0.937 & 0.886 & 0.923 & 0.911 & 0.925 & 0.917 \\
			\midrule
			\multirow{2}{*}{GAP-Net}       & \multirow{2}{*}{IJCV 2023}    & \multirow{2}{*}{4.27}  & \multirow{2}{*}{78.58}   & 33.74 & 33.26 & 34.28 & 41.03   & 31.44 & 32.40 & 32.27 & 30.46 & 33.51 & 30.24 & 33.26 \\
			                               &                               &                        &                          & 0.911 & 0.900 & 0.929 & 0.967   & 0.919 & 0.925 & 0.902 & 0.905 & 0.915 & 0.895 & 0.917 \\
			% \midrule
			% \multirow{2}{*}{ADMM-Net}      & \multirow{2}{*}{ICCV 2019}    & \multirow{2}{*}{4.27}  & \multirow{2}{*}{78.58}   & 34.12 & 33.62 & 35.04 & 41.15   & 31.82 & 32.54 & 32.42 & 30.74 & 33.75 & 30.68 & 33.58 \\
			%                                &                               &                        &                          & 0.918 & 0.902 & 0.931 & 0.966   & 0.922 & 0.924 & 0.896 & 0.907 & 0.915 & 0.895 & 0.918 \\
			% \midrule
			% \multirow{2}{*}{HDNet}         & \multirow{2}{*}{CVPR 2022}    & \multirow{2}{*}{2.37}  & \multirow{2}{*}{154.76}  & 35.14 & 35.67 & 36.03 & 42.30   & 32.69 & 34.46 & 33.67 & 32.48 & 34.89 & 32.38 & 34.97 \\
			%                                &                               &                        &                          & 0.935 & 0.940 & 0.943 & 0.969   & 0.946 & 0.952 & 0.926 & 0.941 & 0.942 & 0.937 & 0.943 \\
			\midrule
			\multirow{2}{*}{MST-L}         & \multirow{2}{*}{CVPR 2022}    & \multirow{2}{*}{2.03}  & \multirow{2}{*}{28.15}   & 35.40 & 35.87 & 36.51 & 42.27   & 32.77 & 34.80 & 33.66 & 32.67 & 35.39 & 32.50 & 35.18 \\
			                               &                               &                        &                          & 0.941 & 0.944 & 0.953 & 0.973   & 0.947 & 0.955 & 0.925 & 0.948 & 0.949 & 0.941 & 0.948 \\
			% \midrule
			% \multirow{2}{*}{MST++}         & \multirow{2}{*}{CVPRW 2022}   & \multirow{2}{*}{1.33}  & \multirow{2}{*}{19.42}   & 35.80 & 36.23 & 37.34 & 42.63   & 33.38 & 35.38 & 34.35 & 33.71 & 36.67 & 33.38 & 35.99 \\
			%                                &                               &                        &                          & 0.943 & 0.947 & 0.957 & 0.973   & 0.952 & 0.957 & 0.934 & 0.953 & 0.953 & 0.945 & 0.951 \\
			\midrule
			\multirow{2}{*}{CST-L}         & \multirow{2}{*}{ECCV 2022}    & \multirow{2}{*}{3.00}  & \multirow{2}{*}{40.01}   & 35.96 & 36.84 & 38.16 & 42.44   & 33.25 & 35.72 & 34.86 & 34.34 & 36.51 & 33.09 & 36.12 \\
			                               &                               &                        &                          & 0.949 & 0.955 & 0.962 & 0.975   & 0.952 & 0.963 & 0.944 & 0.961 & 0.957 & 0.945 & 0.957 \\
			\midrule
			\multirow{2}{*}{S2Transformer} & \multirow{2}{*}{TPAMI 2025}   & \multirow{2}{*}{1.80}  & \multirow{2}{*}{27.21}   & 36.17 & 37.57 & 37.29 & 42.96   & 34.40 & 36.44 & 35.41 & 34.50 & 36.54 & 33.57 & 36.48 \\
			                               &                               &                        &                          & 0.949 & 0.958 & 0.957 & 0.975   & 0.960 & 0.965 & 0.946 & 0.962 & 0.959 & 0.952 & 0.958 \\
			\midrule
			\multirow{2}{*}{BIRNAT}        & \multirow{2}{*}{TPAMI 2023}   & \multirow{2}{*}{4.40}  & \multirow{2}{*}{2122.66} & 36.79 & 37.89 & 40.61 & 46.94 & 35.42 & 35.30 & 36.58 & 33.96 & 39.47 & 32.80 & 37.58 \\
			                               &                               &                        &                          & 0.951 & 0.957 & 0.971 & 0.995 & 0.964 & 0.959 & 0.955 & 0.956 & 0.970 & 0.938 & 0.960 \\
			\midrule
			\multirow{2}{*}{DAUHST-9stg}   & \multirow{2}{*}{NeurIPS 2022} & \multirow{2}{*}{6.15}  & \multirow{2}{*}{79.50}   & 37.25 & 39.02 & 41.05 & 46.15   & 35.80 & 37.08 & 37.57 & 35.10 & 40.02 & 34.59 & 38.36 \\
			                               &                               &                        &                          & 0.958 & 0.967 & 0.971 & 0.983   & 0.969 & 0.970 & 0.963 & 0.966 & 0.970 & 0.956 & 0.967 \\
			\midrule
			\multirow{2}{*}{PADUT-12stg}   & \multirow{2}{*}{ICCV 2023}    & \multirow{2}{*}{5.38}  & \multirow{2}{*}{90.46}   & 37.36 & 40.43 & 42.38 & 46.62   & 36.26 & 37.27 & 37.83 & 35.33 & 40.86 & 34.55 & 38.89 \\
			                               &                               &                        &                          & 0.962 & 0.978 & 0.979 & 0.990   & 0.974 & 0.974 & 0.966 & 0.974 & 0.978 & 0.963 & 0.974 \\
			\midrule
			\multirow{2}{*}{RDLUF-9stg}    & \multirow{2}{*}{CVPR 2023}    & \multirow{2}{*}{1.81}  & \multirow{2}{*}{115.34}  & 37.94 & 40.95 & 43.25 & 47.83 & 37.11 & 37.47 & 38.58 & 35.50 & 41.83 & 35.23 & 39.57 \\
			                               &                               &                        &                          & 0.966 & 0.977 & 0.979 & 0.990 & 0.976 & 0.975 & 0.969 & 0.970 & 0.978 & 0.962 & 0.974 \\
			\midrule
			\multirow{2}{*}{DPU-9stg}      & \multirow{2}{*}{CVPR 2024}    & \multirow{2}{*}{2.85}  & \multirow{2}{*}{49.26}   & \textbf{39.91} & \underline{41.99} & 44.10 & 48.33 & 38.07 & 38.58 & 39.13 & 36.90 & 42.88 & 36.36 & 40.52 \\
			                               &                               &                        &                          & 0.968 & 0.981 & 0.980 & 0.990 & 0.978 & 0.978 & 0.971 & 0.975 & 0.981 & 0.970 & 0.977 \\
			\midrule
			\multirow{2}{*}{SSR-L}         & \multirow{2}{*}{CVPR 2024}    & \multirow{2}{*}{5.18}  & \multirow{2}{*}{78.93}   & 39.07 & \textbf{42.04} & \textbf{44.49} & \textbf{48.80} & 38.64 & 38.50 & 39.16 & 36.96 & 43.12 & 36.08 & 40.69 \\
			                               &                               &                        &                          & 0.970 & 0.981 & 0.980 & 0.990 & 0.980 & 0.978 & 0.971 & 0.976 & 0.982 & 0.968 & 0.978 \\
			\midrule
			\multirow{2}{*}{LADE-DUN-9stg} & \multirow{2}{*}{ECCV 2024}    & \multirow{2}{*}{2.78}  & \multirow{2}{*}{88.68}   & 38.08 & 41.84 & 43.77 & 47.99 & 37.97 & 38.30 & 38.82 & 36.15 & 42.53 & 35.48 & 40.09 \\
			                               &                               &                        &                          & 0.969 & \underline{0.982} & \textbf{0.983} & \underline{0.993} & 0.980 & \textbf{0.980} & 0.973 & 0.979 & \underline{0.984} & 0.970 & \underline{0.979} \\
			\midrule
			\multirow{2}{*}{MiJUN-9stg}    & \multirow{2}{*}{AAAI 2025}    & \multirow{2}{*}{0.56}  & \multirow{2}{*}{73.67}   & 39.26 & 41.78 & \underline{44.31} & 48.53 & \underline{39.30} & 38.22 & \textbf{41.00} & 36.72 & \textbf{43.84} & 35.56 & \underline{40.86} \\
			                               &                               &                        &                          & \textbf{0.973} & \textbf{0.983} & \textbf{0.983} & \textbf{0.994} & \textbf{0.985} & \underline{0.979} & \textbf{0.983} & 0.978 & \textbf{0.985} & 0.967 & \textbf{0.982} \\
			\midrule
			\rowcolor{rouse}
			                               &                               &                        &                          & 38.39 & 39.66 & 42.07 & 47.43 & 38.08 & 37.10 & 38.47 & 36.50 & 41.26 & 35.40 & 39.44 \\
			\rowcolor{rouse}
			\multirow{-2}{*}{LRDUN-3stg}   & \multirow{-2}{*}{Ours}        & \multirow{-2}{*}{0.69} & \multirow{-2}{*}{10.26}  & 0.964 & 0.969 & 0.976 & 0.988 & 0.977 & 0.971 & 0.970 & 0.966 & 0.976 & 0.961 & 0.972 \\
			\midrule
			\rowcolor{rouse}
			                               &                               &                        &                          & \underline{39.48} & 40.44 & 42.96 & 48.11 & 38.76 & 37.91 & 39.62 & 37.52 & 42.59 & 35.59 & 40.30 \\
			\rowcolor{rouse}
			\multirow{-2}{*}{LRDUN-6stg}   & \multirow{-2}{*}{Ours}        & \multirow{-2}{*}{1.37} & \multirow{-2}{*}{20.45}  & 0.970 & 0.973 & 0.979 & 0.992 & 0.981 & 0.974 & 0.976 & 0.972 & 0.981 & 0.965 & 0.976 \\
			\midrule
			\rowcolor{rouse}
			                               &                               &                        &                          & 39.45 & 40.97 & 43.63 & \underline{48.76} & \textbf{39.36} & \textbf{38.99} & \underline{40.24} & \textbf{38.28} & \underline{43.39} & \textbf{36.52} & \textbf{40.96} \\
			\rowcolor{rouse}
			\multirow{-2}{*}{LRDUN-9stg}   & \multirow{-2}{*}{Ours}        & \multirow{-2}{*}{2.04} & \multirow{-2}{*}{30.58}  & \textbf{0.973} & 0.979 & \underline{0.982} & \underline{0.993} & \underline{0.984} & \textbf{0.980} & \underline{0.979} & \textbf{0.981} & \textbf{0.985} & \textbf{0.973} & \textbf{0.982} \\
			\midrule
			\rowcolor{rouse}
			                               &                               &                        &                          & 39.34 & 40.68 & 43.39 & 48.57 & 39.14 & \underline{38.78} & 40.02 & \underline{37.96} & 43.12 & \underline{36.49} & 40.75  \\
			\rowcolor{rouse}
			\multirow{-2}{*}{LRDUN-9stg*}  & \multirow{-2}{*}{Ours}        & \multirow{-2}{*}{0.25} & \multirow{-2}{*}{30.58}  & \underline{0.971} & 0.975 & 0.981 & \underline{0.993} & 0.982 & \underline{0.979} & 0.976 & \underline{0.980} & \underline{0.984} & \underline{0.972} & \underline{0.979} \\
			\bottomrule[0.2em]
		\end{tabular}
	} % 结束 resizebox
	\vspace{-5mm}
\end{table*}

\subsection{Overall Architecture}
\label{subsec:over}
The overall architecture of the proposed LRDUN is illustrated in \figref{fig:arch}(a).
The network unfolds the alternating optimization into $N$ iterative stages, taking the compressed measurement $y$ and sensing matrix $\Phi$ as inputs. Initially, Init Net $\ten{A}$ and Init Net $\mat{E}$ generate the initial feature representations $\ten{A}_{feat}^{0}$ and $\mat{E}_{feat}^{0}$, which are iteratively refined across $N$ stages. 
% At each stage, the network alternates between updating the featurized data-fidelity terms under the GFUM (\figref{fig:arch}(b)) introduced in \secref{subsec:GFUM} and the learnable ProxyNets (\figref{fig:arch}(c)-(d)). After iterating through all $N$ stages, the network outputs the final spectral basis $E^{N}$ and spatial subspace image $\mathcal{A}^{N}$.
% The reconstructed HSI is then obtained via $\ten{X} = \ten{A}^{N} \times_3 \mat{E}^{N}$.
Each stage alternates between solving the E-problem and the A-problem, where each subproblem update involves: (1) applying a data-fidelity feature term via the GFUM (\figref{fig:arch}(b)) introduced in \secref{subsec:GFUM}, and (2) refining the result with a learnable ProxyNet (\figref{fig:arch}(c)-(d)). After $N$ iterations, the network outputs the final spectral basis $\mat{E}^{N}$ and spatial subspace image $\ten{A}^{N}$. The reconstructed HSI is then obtained via $\ten{X} = \ten{A}^{N} \times_3 \mat{E}^{N}$.

\subsection{The Structure of ProxyNet}
\label{subsec:proxynet}

% As shown in Fig. 2(c), $\mathrm{ProxyNet}_{\mathrm{E}}$ adopts a lightweight 1D convolutional architecture operating along the spectral dimension, using stacked 1D convolutions with GELU activation and residual shortcuts to refine $\mat{E}_{feat}$ and capture local spectral correlations. Remarkable, we further apply a QR decomposition  to ensure the column orthonormality of $\mat{E}$, i.e., $\mat{E}^T \mat{E} = \mat{I}$. 
As shown in \figref{fig:arch}(c), $\mathrm{ProxyNet}_{\mathrm{E}}$ is a lightweight 1D architecture. It refines $\mat{E}_{feat}$ using stacked 1D convolutions with GELU activation and residual shortcuts to capture local spectral correlations. Crucially, we apply a QR decomposition to the physical component $\mat{E}$ to enforce its column orthonormality ($\mat{E}^T \mat{E} = \mat{I}$). In \figref{fig:arch}(d), $\mathrm{ProxyNet}_{\mathrm{A}}$ adopts a U-Net architecture to model spatial dependencies within the subspace features. The encoder-decoder structure employs symmetrical Spatial Conv Attention Blocks (SCAB) and Down/Up-sampling Blocks. 
% The SCAB leverages a Conv-like Attention architecture, as illustrated in \figref{fig:sab_detail}. 
The SCAB aims to efficiently model long-range dependencies and selectively enhance features via a Conv-like Attention mechanism, as illustrated in \figref{fig:sab_detail}. The details of ProxyNet architectures are provided in the \textbf{supplementary material}.

\section{Experiments}

\subsection{Datasets}
We evaluate the proposed LRDUN on both simulated and real hyperspectral datasets.
For simulated experiments, we adopt two widely used benchmarks: \textbf{CAVE}~\cite{parkMultispectralImagingUsing2007} and \textbf{KAIST}~\cite{choiHighqualityHyperspectralReconstruction2017}.
Following TSA-Net~\cite{mengEndtoEndLowCost2020}, a real coded aperture mask of size $256 \times 256$ is employed for simulation.
The model is trained on the CAVE dataset and tested on ten scenes from KAIST, each cropped to $256 \times 256$ pixels for fair comparison.
For real-world evaluation, we use five compressive measurements (spatial size $660 \times 714$) captured by the CASSI system described in~\cite{mengEndtoEndLowCost2020}.

\subsection{Implementation Details}
Our LRDUN is implemented in PyTorch and trained on a single NVIDIA RTX 4090 GPU.
A multi-stage RMSE loss \cite{zhangDualPriorUnfolding2024} supervises reconstruction at each stage.
The model is optimized with Adam using an initial learning rate of $4 \times 10^{-4}$ under a cosine annealing schedule, 
for 300 epochs with a batch size of 2. We set the number of stages $N=3, 6, 9$ for different model sizes, with
physical rank $k=11$ and the feature dimension $C=16$ in all experiments.

\subsection{Comparison Methods}
We compare LRDUN with a broad range of state-of-the-art (SOTA) approaches, categorized into four groups:

\noindent\textbf{Model-based methods:} TwIST~\cite{bioucas-diasNewTwISTTwoStep2007},
GAP-TV~\cite{yuanGeneralizedAlternatingProjection2016} and
DeSCI~\cite{liuRankMinimizationSnapshot2019}.
\noindent\textbf{E2E methods:}
$\lambda$-Net~\cite{miaoLambdaNetReconstructHyperspectral2019},
% HDNet~\cite{huHDNetHighresolutionDualdomain2022},
TSA-Net~\cite{mengEndtoEndLowCost2020},
BIRNAT~\cite{chengRecurrentNeuralNetworks2023},
MST~\cite{caiMaskguidedSpectralwiseTransformer2022},
% MST++~\cite{caiMSTMultistageSpectralwise2022},
CST~\cite{caiCoarsetoFineSparseTransformer2022}
and S2Transformer~\cite{wangS2TransformerMaskAwareHyperspectral2025}.
\noindent\textbf{PnP methods:} DIP-HSI \cite{zhengDeepPlugandplayPriors2021}.
\noindent\textbf{DUNs:}
% HSSP~\cite{wangHyperspectralImageReconstruction2019},
% ADMM-Net~\cite{maDeepTensorADMMNet2019},
DGSMP~\cite{huangDeepGaussianScale2021},
GAP-Net~\cite{mengDeepUnfoldingSnapshot2023},
DAUHST~\cite{caiDegradationawareUnfoldingHalfshuffle2022},
PADUT~\cite{liPixelAdaptiveDeep2023},
RDLUF~\cite{dongResidualDegradationLearning2023},
DPU~\cite{zhangDualPriorUnfolding2024},
SSR~\cite{zhangImprovingSpectralSnapshot2024},
LADE-DUN~\cite{wuLatentDiffusionPrior2025}
and MiJUN~\cite{qinDetailMattersMambaInspired2025}.

% \subsection{Evaluation Metrics}
% We quantitatively assess reconstruction performance using the \textbf{Peak Signal-to-Noise Ratio (PSNR)} and the \textbf{Structural Similarity Index (SSIM)}, which respectively measure pixel-level fidelity and perceptual consistency between the reconstructed and reference hyperspectral images.

\begin{figure*}[t]
	\centering
	\includegraphics[width=0.98\textwidth]{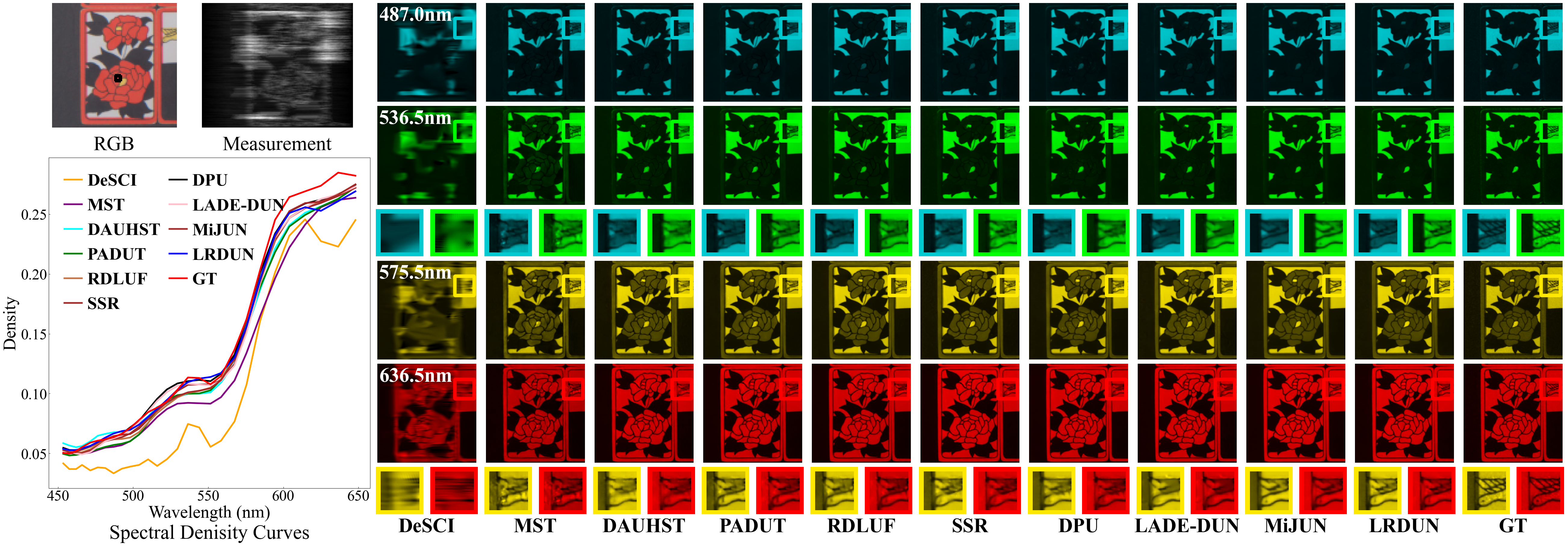}
	\vspace{-3mm}
	\caption{
		Reconstructed results of the simulated Scene~7, showing 4 out of 28 spectral channels obtained by state-of-the-art methods.
		One representative region is selected for spectral analysis.
	}
	\label{fig:simulation_visual}
\end{figure*}

\begin{figure*}[t]
	\centering
	\includegraphics[width=0.98\textwidth]{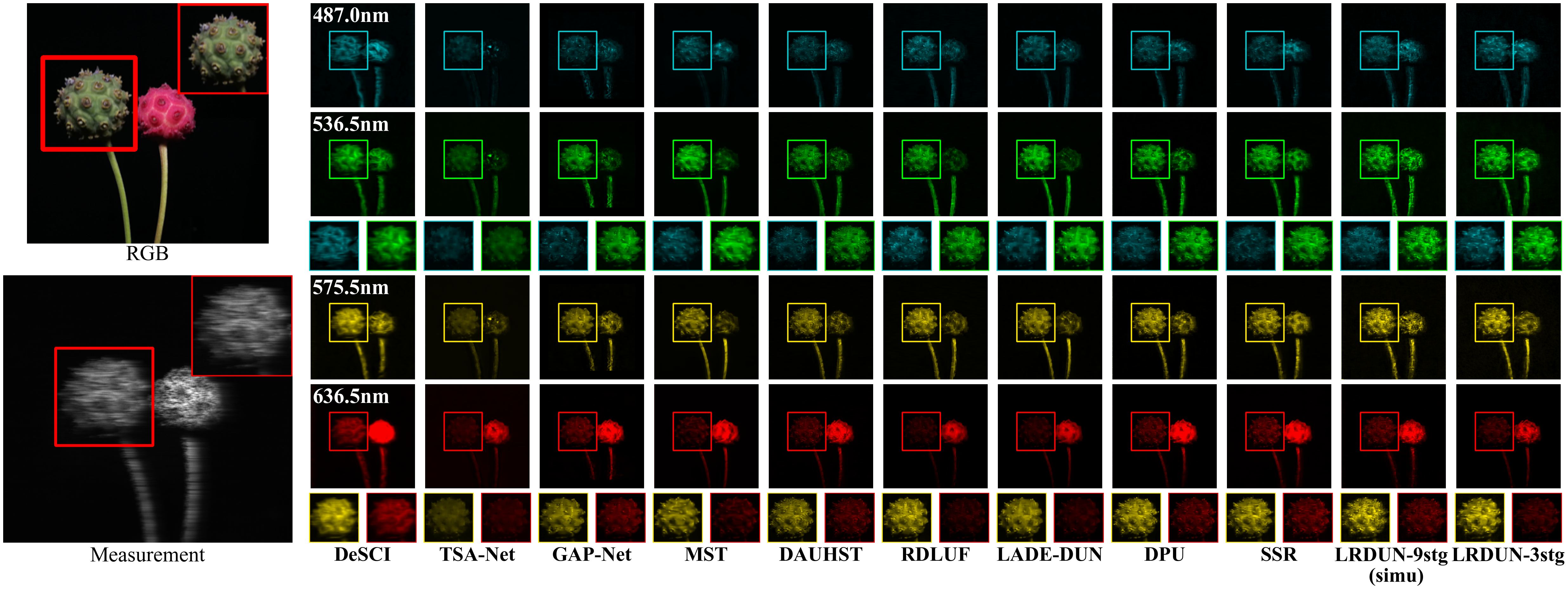}
	\vspace{-3mm}
	\caption{
		Reconstructed results of real-world Scene~4, displaying 4 out of 28 spectral channels. LRDUN-9stg (simu) denotes the model obtained directly from the Simulation Experiment.
	}
	\label{fig:real_visual}
	\vspace{-4mm}
\end{figure*}

\subsection{Simulation Results}

\noindent\textbf{Quantitative Results.}
\tabref{tab:simu} presents the quantitative comparison across ten simulated scenes. The proposed LRDUN consistently outperforms all competing approaches in both PSNR and SSIM, confirming its robustness and generalization ability. Specifically, LRDUN-9stg achieves the highest average PSNR of 40.96 dB, surpassing the SOTA DUNs.
It is worth noting that both RDLUF-9stg and MiJUN-9stg employ parameter-sharing across stages, leading to smaller parameter counts but significantly larger FLOPs. In contrast, our LRDUN strikes a more balanced trade-off, achieving superior reconstruction quality with moderate computational complexity. Moreover, the LRDUN-9stg* adopts the same parameter-sharing strategy as RDLUF and MiJUN. Despite its greatly reduced parameter size, it maintains nearly identical accuracy, verifying the efficiency and scalability of the proposed architecture.

% Table~\ref{tab:simu} reports the quantitative results on 10 simulated scenes.
% Our LRDUN consistently achieves superior performance across all datasets and metrics, verifying the effectiveness and robustness of the proposed framework.
% Compared with Transformer-based models such as MST-L and CST-L, LRDUN attains an average gain of over \textbf{0.8 dB} in PSNR while requiring fewer GFLOPs.
% Even the compact version achieves comparable accuracy at a fraction of the computational cost.
% Figure~\ref{fig:ssim_gflops} further illustrates the trade-off between SSIM and GFLOPs, where LRDUN achieves the best balance between performance and efficiency.

\noindent\textbf{Visual Results.}
Following~\cite{mengEndtoEndLowCost2020}, reconstructed HSIs are visualized in RGB using CIE color mapping. As shown in \figref{fig:simulation_visual}, competing methods struggle to recover intricate textures and boundaries with noticeable artifacts and oversmoothing. 
% suffer from artifacts and oversmoothing, failing to recover fine-grained textures and boundaries. 
% In contrast, LRDUN consistently yields clearer spatial structures and sharper edges. 
In contrast, LRDUN consistently delivers better visual results with clearer spatial structures and sharper edges.
Such superior fidelity is largely credited to the SCAB module, which leverages a substantial $11\times11$ receptive field to model the long-range dependencies vital for recovering intricate spatial details.

\subsection{Real-World Results}
We further evaluate our method on real CASSI measurements, where ground-truth HSIs are unavailable. Following prior work \cite{mengEndtoEndLowCost2020}, we retrained an LRDUN-3stg model. Crucially, we also directly applied the LRDUN-9stg (simu) model from the Simulation Experiment without any fine-tuning. 
As shown in \figref{fig:real_visual}, 
% while competing methods suffer from significant texture loss, LRDUN effectively preserves fine spatial structures and spectral consistency. 
whereas competing methods exhibit substantial texture degradation, LRDUN successfully preserves fine spatial structures and maintains spectral consistency.
% The robust performance of the non-fine-tuned LRDUN-9stg (simu) on complex, noisy real data validates its strong generalization. 
Notably, the robust performance of the non-fine-tuned LRDUN-9stg on real-world data underscores its exceptional generalization.
This robustness stems from its principled low-rank reformulation, which effectively alleviates the severe ill-posedness of the reconstruction task.

\subsection{Ablation Study}
\label{subsec:feat_visual}
\begin{table}[tbp]
	\centering
	% \scriptsize
	\caption{Ablation of different LR embedding strategies.}
	\vspace{-3mm}
	\label{tab:ablation_lre}
	\resizebox{\columnwidth}{!}{%
		\begin{tabular}{c c c c c c c}
			\toprule
			Metric     & Baseline-1 & w. NNL & w. TSVD & w. SP-1 & w. SP-2 & LRDUN          \\
			\midrule 
			PSNR (dB)  & 38.16      & 37.66  & 37.93   & 38.33  & 38.52 & \textbf{39.44} \\
			Params (M) & 1.87       & 1.87   & 1.87    & 1.87   & 1.87  & \textbf{0.69}  \\
			FLOPs (G)  & 26.95      & 26.95  & 26.95   & 26.95  & 26.95 & \textbf{10.26} \\
			\bottomrule
		\end{tabular}%
	}
    \vspace{-3mm}
\end{table}

\noindent\textbf{LR Embedding Strategy.} 
We first analyze the impact of different LR embedding strategies. A 3-stage model that directly processes the full HSI using only $\mathrm{ProxyNet}_{\mathrm{A}}$ is taken as Baseline-1.  As shown in \tabref{tab:ablation_lre}, Baseline-1 suffers from prohibitive computational overhead and limited performance. 
We compare Baseline-1 with several variants: applying a nuclear norm loss (NNL), integrating a truncated SVD (TSVD) layer, adopting the SP method \cite{heInterpretableFlexibleFusion2024} and our proposed LRDUN. The NNL and TSVD strategies suffer from training instability, resulting in performance degradation. Although the SP method provides a slight performance improvement, they rely on the burdensome full-HSI reconstruction from Baseline-1 and fail to alleviate the underlying computational strain. In contrast, LRDUN circumvents this bottleneck by decomposing the sensing model into efficiently solvable subproblems. It not only significantly boosts reconstruction accuracy (39.44 dB) but also drastically reduces computational complexity by 61.9\%.

% We establish Baseline-1 as a 3-stage (3stg) model that processes the full HSI using only $\mathrm{ProxyNet}_{\mathrm{A}}$. To validate the effectiveness of different LR embedding strategies, we compare Baseline-1 against several variants: applying a nuclear norm loss (NNL), integrating a truncated SVD (TSVD) layer for intermediate features, the SP method \cite{heInterpretableFlexibleFusion2024}, and our proposed LRDUN.As shown in Table~\ref{tab:ablation_lre}, Baseline-1 suffers from prohibitive computational overhead and limited performance. While integrating different LR embedding strategies enhances reconstruction performance to varying degrees, their efficacy differs significantly. Directly applying the NNL yields the least improvement, indicating that a simple LR prior is insufficient for effective performance gains. The TSVD layer provides a moderate boost, but its effectiveness is constrained by its non-adaptive nature. The SP method further improves performance; however, as it relies on the full HSI reconstruction from Baseline-1, it still incurs a substantial computational burden. In contrast, our proposed LRDUN reformulates the problem by alternately optimizing the A and E subproblems. This approach not only significantly boosts reconstruction accuracy but also drastically reduces computational complexity.

\begin{figure}[tbp]
    \centering
    \scriptsize
    \includegraphics[width=0.47\textwidth]{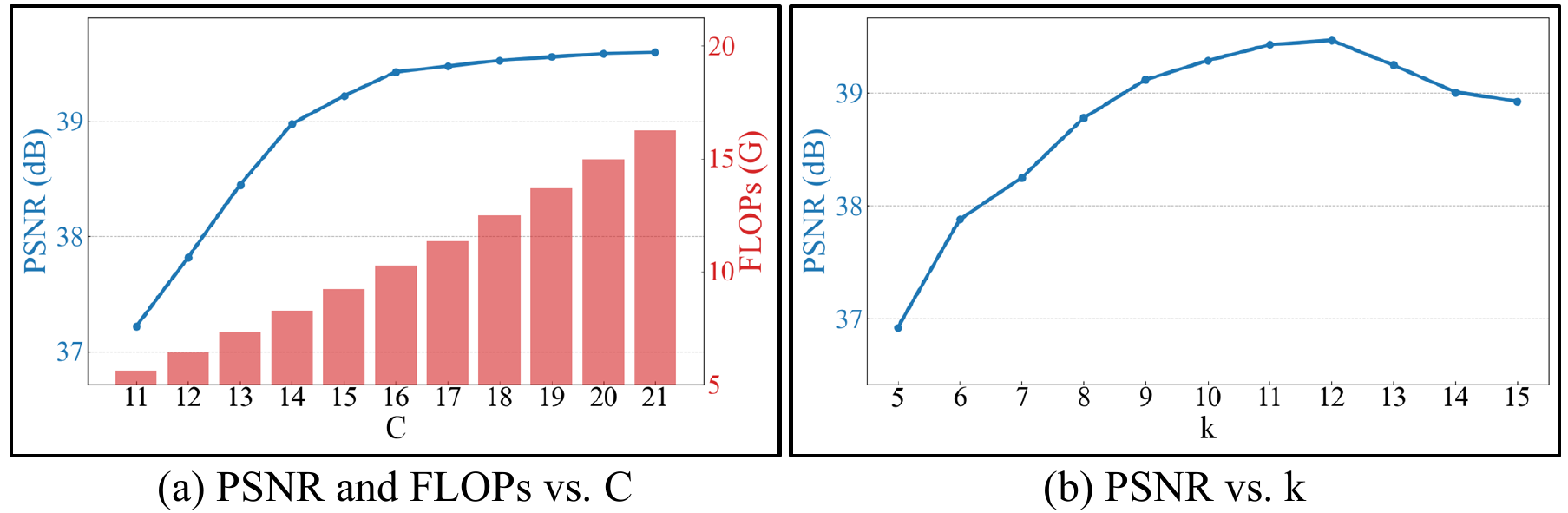}
    \vspace{-3mm}
    \caption{PSNR and FLOPs under different settings of $C$ and $k$.}
    \label{fig:ab_Ck}
    \vspace{-6mm}
\end{figure}

\begin{figure}[tbp]
    \centering
    \includegraphics[width=0.47\textwidth]{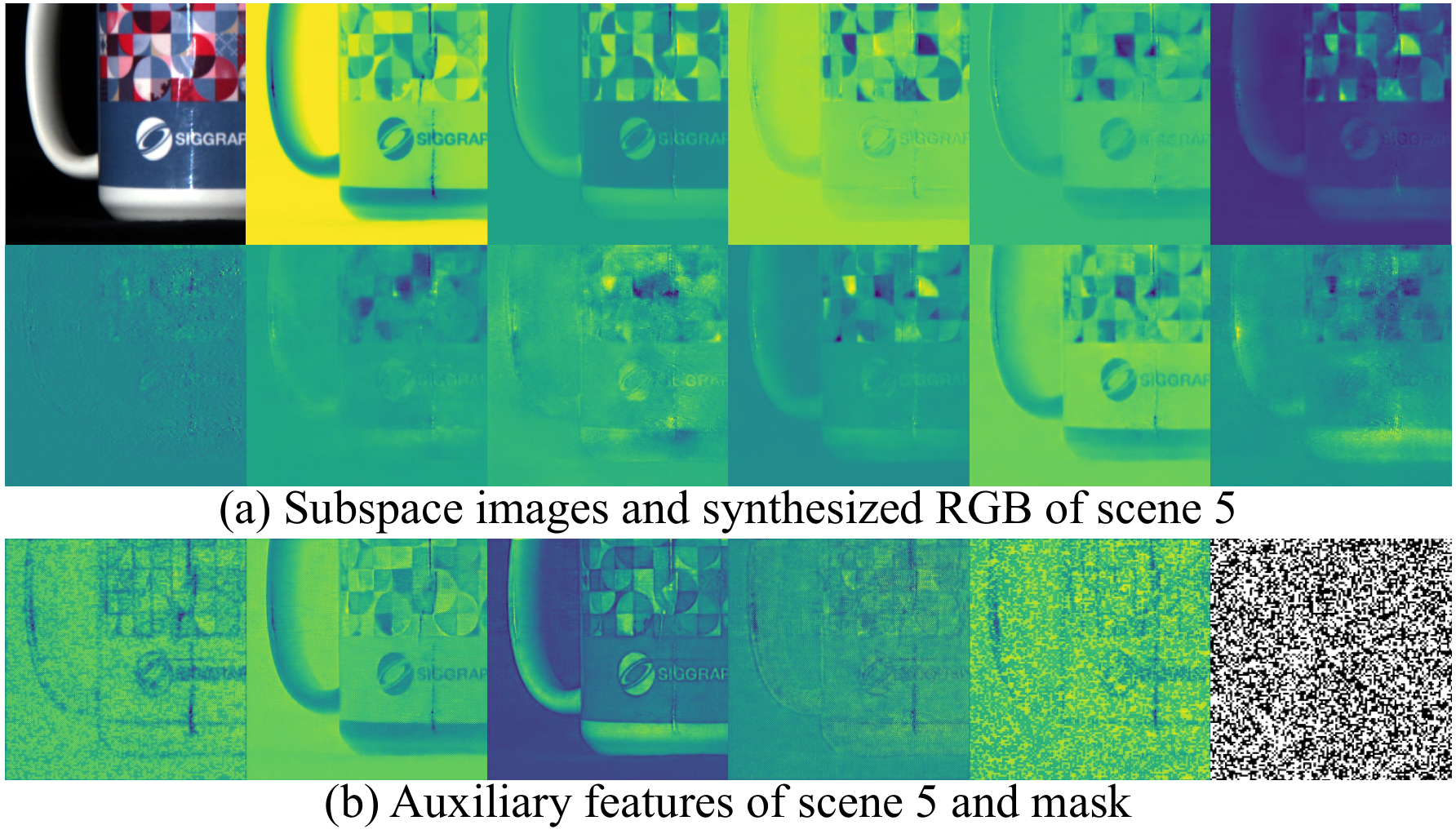}
    \vspace{-3mm}
    \caption{Visualization of the GFUM.}
    \label{fig:GFUM_visual}
    \vspace{-3mm}
\end{figure}

\noindent\textbf{Effectiveness of GFUM.} 
We next examine the effect of the GFUM by varying the feature dimension $C$ within the ProxyNets. The case $C=k$ corresponds to the absence of GFUM (i.e., constraining the feature dimension to the physical rank), whereas $C>k$ activates it. As shown in \figref{fig:ab_Ck}(a), increasing $C$ consistently enhances reconstruction quality, demonstrating that GFUM effectively expands the representational capacity of ProxyNets. However, larger feature dimensions also increase FLOPs. Balancing accuracy and efficiency, we set $C=16$ (for $k=11$) as the default configuration. To further understand GFUM's behavior, \figref{fig:GFUM_visual} visualizes the physical and auxiliary components from $\mathrm{ProxyNet}_{\mathrm{A}}$. The physical part primarily captures structural and semantic content (e.g., object contours and logos), while the auxiliary part encodes complementary, spatially-varying information such as high-frequency textures, background-foreground separation, and mask-aware cues. This complementary information flow validates the role of the auxiliary component in improving fine-detail recovery and noise suppression.

\noindent\textbf{Rank Selection.} 
Keeping $C=16$ fixed, we vary the physical rank $k$ to assess its influence on performance. As illustrated in \figref{fig:ab_Ck}(b), the reconstruction performance increases with $k$ up to a point, beyond which the benefit saturates or even declines. An excessively small $k$ fails to model the intrinsic diversity of HSI spectra, while an overly large $k$ compromises the benefit of auxiliary features by diminishing the auxiliary component's dimensionality ($C-k$). Considering the trade-off, we choose $k=11$ as the optimal rank.

% the auxiliary component's dimensionality ($C-k$), limiting ProxyNet's representational capacity. To balance accuracy and efficiency, we choose $k=11$ as our final configuration.
% k fails to model the intrinsic diversity of HSI spectraA larger $k$ reduces the auxiliary component's dimensionality ($C-k$), limiting ProxyNet's representational capacity. To balance accuracy and efficiency, we choose $k=11$ as our final configuration.

% We fix the feature dimension at $C=16$ and vary the physical rank $k$ to analyze its impact on performance and computational cost. As shown in \figref{fig:ab_Ck}(b), reconstruction performance generally improves with a larger $k$. However, an excessively large $k$ (which reduces the dimensionality of the auxiliary component, $C-k$) begins to limit the ProxyNet's representational capacity, leading to diminishing returns. To achieve the best trade-off between performance and efficiency, we select $k=11$ as our final setting.

\begin{table}[tbp]
	\centering
	% \scriptsize
	\caption{Attention Comparisons.}
	\vspace{-3mm}
	\label{tab:ab_attn}
	\resizebox{\columnwidth}{!}{%
		\begin{tabular}{c c c c c c}
			\toprule
			Metric     & Baseline-2 & W-MSpaA & W-MSpeA & HS-MSpaA & SCAB  \\
			\midrule
			PSNR (dB)  & 37.48      & 39.13   & 39.02   & 39.31    & 39.44 \\
			Params (M) & 0.56       & 0.74    & 0.74    & 0.74     & 0.69  \\
			FLOPs (G)  & 7.55       & 9.73    & 9.70    & 9.89     & 10.26 \\
			\bottomrule
		\end{tabular}%
	}
	\vspace{-6mm}
\end{table}

\noindent\textbf{Attention Comparison.} 
Finally, we assess different attention mechanisms. 
We define Baseline-2 as a simplified LRDUN-3stg with the SCAB module removed. We compare W-MSpaA \cite{liuSwinTransformerHierarchical2021}, W-MSpeA \cite{liPixelAdaptiveDeep2023,zhangImprovingSpectralSnapshot2024}, HS-MSA \cite{caiDegradationawareUnfoldingHalfshuffle2022}, and our proposed SCAB. 
As reported in \tabref{tab:ab_attn}, SCAB achieves the best performance with moderate computational cost, highlighting its ability to effectively model long-range dependencies and enhancing feature representations.

\section{Conclusion}

% Existing DUNs suffer from severe ill-posedness and computational inefficiency stemming from their reliance on the full-HSI imaging model. To address these challenges, 
We fundamentally reformulate the SCI problem by deriving two novel imaging models that allow the network to recover compact spectral basis and subspace images instead of the high-dimensional data cube. This paradigm shift significantly mitigates the ill-posedness while embedding strong physical priors. Building on these models, we propose LRDUN that alternately optimizes the spectral and spatial components within an unfolded PGD framework. Furthermore, we introduce a  GFUM  to decouple the physical rank from the feature dimensionality, substantially enhancing representational capacity. Extensive experiments demonstrate that LRDUN achieves SOTA  reconstruction quality with significantly reduced computational costs. We believe LRDUN establishes a new paradigm for physics-informed, interpretable, and efficient SCI reconstruction.

\section*{Acknowledgments} 
This work was supported in part by the National Key Research and Development Program of China under Grant 2022YFB3903501 and in part by the National Natural Science Foundation of China under Grant 42271370.

{
	\small
	\bibliographystyle{reference/ieeenat_fullname}
	\bibliography{reference/main.bib}
}


% \input{supplementary/supp.tex}
% \input{supplementary/supp.tex}



\section*{Supplementary material}

# LRDUN: A Low-Rank Deep Unfolding Network for Efficient Spectral Compressive Imaging

## Supplementary Material

## 1. QR Retraction in $\text{ProxyNet}_E$

The low-rank decomposition of a tensor $\mathcal{X}$ into $\mathcal{X} = \mathcal{A} \times_3 \mathbf{E}$, where $\mathcal{A} \in \mathbb{R}^{H \times W \times k}$ and $\mathbf{E} \in \mathbb{R}^{B \times k}$, inherently suffers from scale ambiguity (or indeterminacy). Specifically, for any non-zero scalar $a \in \mathbb{R}$, the pair $\mathbf{E}' = a\mathbf{E}$ and $\mathcal{A}' = \mathcal{A}/a$ constitutes an equally valid decomposition. To resolve this indeterminacy and stabilize the optimization process, we enforce an orthogonality constraint on the spectral basis $\mathbf{E}$, thereby requiring $\mathbf{E}$ to reside on the Stiefel manifold ($\mathbf{E}^T \mathbf{E} = \mathbf{I}$). For the resulting constrained optimization problem, we adopt the widely-used QR retraction method [6] to project the updated basis $\mathbf{E}$ back onto the Stiefel manifold after each iteration. Mathematically, this involves first performing a QR decomposition on the updated $\mathbf{E}$:

$$\mathbf{E} = \mathbf{QR},$$

where $\mathbf{Q} \in \mathbb{R}^{B \times k}$ is a column-orthogonal matrix and $\mathbf{R} \in \mathbb{R}^{k \times k}$ is an upper triangular matrix. The retracted spectral basis is then set as $\mathbf{E} = \mathbf{Q}$.

To validate the efficacy of QR Retraction, we compare it against a constraint-free Baseline-3 and a soft orthogonal loss ($\mathcal{L} = \|\mathbf{E}^T \mathbf{E} - \mathbf{I}\|_F^2$). As shown in Table S1, the baseline fails to converge due to numerical instability. The orthogonal loss, acting merely as a soft constraint, yields suboptimal performance as it cannot enforce strict orthogonality. In contrast, QR Retraction guarantees strict orthogonality via manifold optimization, significantly improving both stability and reconstruction quality.

Table S1. Ablation Study on QR Retraction.

| Metric | Baseline-3 | Orth. Loss | QR Retraction |
| --- | --- | --- | --- |
| PSNR (dB) | not converged | 38.18 | 39.43 |
| SSIM | not converged | 0.963 | 0.972 |

## 2. SCAB in $\text{ProxyNet}_A$

Convolutional attention mechanisms have proven highly successful in various computer vision tasks, such as image classification [2, 4], semantic segmentation [1, 3], and image resolution [5], owing to their efficient and effective encoding of contextual information. In our LRDUN framework, the $\text{ProxyNet}_A$ module employs a SCAB to efficiently model long-range spatial dependencies within the subspace features. As illustrated in Fig. 3 of the main paper, the SCAB architecture processes input features through two parallel pathways. The first is the **Attention Path**, where the input feature is processed by a `Conv2D` $1 \times 1$ layer, followed by a large-kernel `DWConv2D` $11 \times 11$ ($11 \times 11$ Depth-wise Convolution) to capture features with long-range spatial dependencies. The resulting output is passed through a `GELU` activation to generate the spatial attention map. Concurrently, the **Value Path** processes the input feature with a separate `Conv2D` $1 \times 1$ layer to produce the value representation. The two representations are then fused via element-wise multiplication ($\odot$), allowing the long-range contextual information from the attention path to selectively modulate the features in the value path. This fused feature is then passed through a final `Conv2D` $1 \times 1$ layer, aggregated with the initial input via a residual connection, and finally refined by a ConvFFN block. This design provides an efficient mechanism for modeling long-range dependencies and enhancing feature representations, which is validated by its superior performance (39.44 dB) against other attention mechanisms in our ablation study (Table 3 of the main paper).

## 3. More Visual Comparison Results

To further validate the effectiveness of our proposed LRDUN method, we provide additional visual comparison results on both real and simulated datasets. Figures S1 to S4 present reconstructed results of real-world scenes, showcasing 4 out of 28 spectral channels for each scene. Figures S5 to S8 display reconstructed results of simulated scenes obtained by various DUNs, along with spectral analysis of representative regions. These visualizations further demonstrate the superior reconstruction quality and spectral fidelity achieved by our LRDUN approach.

## References

[1] Han Cai, Junyan Li, Muyan Hu, Chuang Gan, and Song Han. Efficientvit: Lightweight multi-scale attention for high-resolution dense prediction. In *Proceedings of the IEEE/CVF International Conference on Computer Vision*, pages 17302–17313, 2023. 1

[2] Yinpeng Chen, Xiyang Dai, Mengchen Liu, Dongdong Chen, Lu Yuan, and Zicheng Liu. Dynamic convolution: Attention over convolution kernels. In *Proceedings of the IEEE/CVF Conference on Computer Vision and Pattern Recognition*, pages 11030–11039, 2020. 1

[3] Meng-Hao Guo, Cheng-Ze Lu, Qibin Hou, Zhengning Liu, Ming-Ming Cheng, and Shi-Min Hu. Segnext: Rethinking convolutional attention design for semantic segmentation. *Ad-*

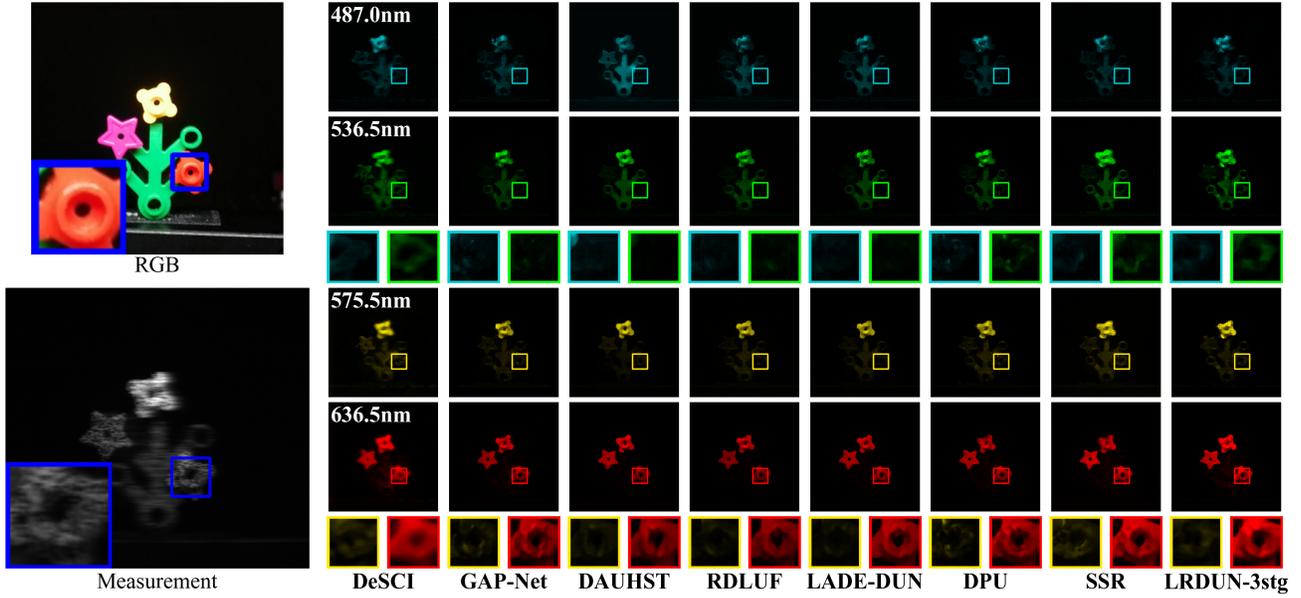

Figure S1. Reconstructed results of real-world Scene 1, displaying 4 out of 28 spectral channels.

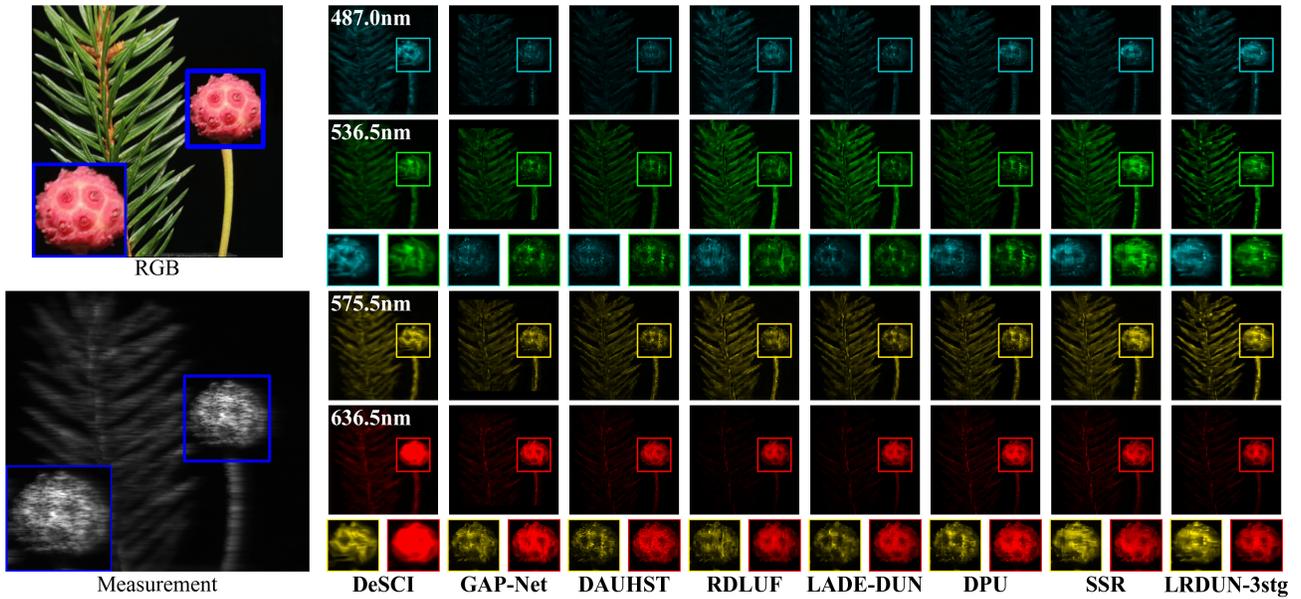

Figure S2. Reconstructed results of real-world Scene 2, displaying 4 out of 28 spectral channels.

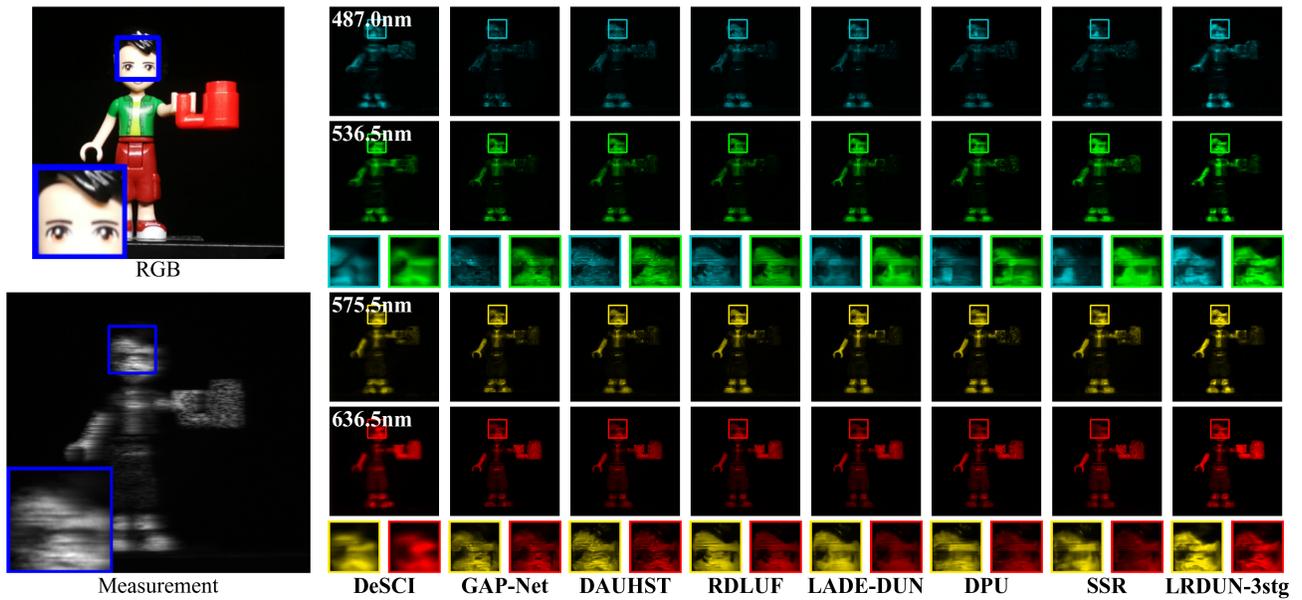

Figure S3. Reconstructed results of real-world Scene 3, displaying 4 out of 28 spectral channels.

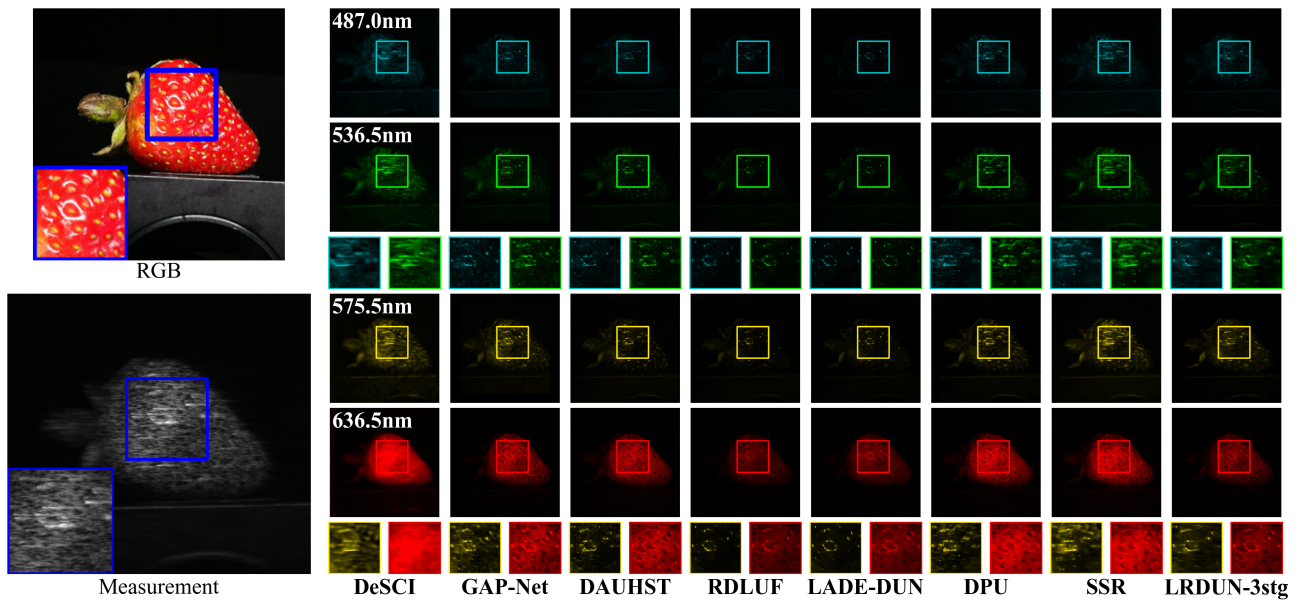

Figure S4. Reconstructed results of real-world Scene 5, displaying 4 out of 28 spectral channels.

3

 % 插入所有页面

\end{document}